 \documentclass[final,3p,times,twocolumn]{elsarticle}

\usepackage{amsmath, amssymb}
\usepackage{booktabs}
\usepackage{multirow}
\usepackage{hyperref}
\usepackage{algorithm}
\usepackage{algpseudocode}
\usepackage{subcaption}
\usepackage{svg}
\journal{Neurocomputing}

\begin{document}

\begin{frontmatter}

\title{Symbolic Branch Networks: Tree-Inherited Neural Models for Interpretable Multiclass Classification}

\author{Dalia N\"oth-Rodr\'iguez}
\ead{dalia.noeth-rodriguez@gmx.de}

\affiliation{organization={Independent Researcher},
            country={Germany}}

\begin{abstract}
Symbolic Branch Networks (SBNs) are neural models whose architecture is
derived directly from an ensemble of decision trees. Each root-to-parent-of-leaf
decision path is mapped to a hidden neuron, while the matrices $W_{1}$
(feature-to-branch) and $W_{2}$ (branch-to-class) encode the structural
relationships induced by the teacher ensemble. This provides an explicit
correspondence between teacher branches and neural components, enabling
feature-level and branch-level interpretation while retaining the flexibility
of gradient-based optimization.
The primary configuration studied in this work trains the teacher-derived
feature-to-branch matrix $W_{1}$ through gradient-based learning while
keeping the branch-to-class matrix $W_{2}$ fixed; this configuration is
denoted SBN$_{1,0}$. Across 29 multiclass tabular datasets, this
configuration improves upon the teacher ensemble on 28 of 29 datasets,
increasing mean accuracy from 84.0\% to 87.9\%. For reference, XGBoost
and TabPFN achieve mean accuracies of 87.3\% and 91.3\%, respectively, on
the same datasets. A one-sided Wilcoxon signed-rank test on dataset-level
mean accuracies indicates a statistically significant improvement of
SBN$_{1,0}$ over the teacher ensemble ($p=4.71\times10^{-7}$).
Beyond predictive performance, SBN provides an explicit structural basis for
interpretation. Global feature importance can be derived from the
teacher-induced feature-to-branch connectivity, while individual predictions
can be traced to selected branch-neurons, their associated features, and
their class contributions. An analysis across ten independently constructed
teacher ensembles shows consistent predictions for a selected instance but
only partial stability in its local feature rankings. We also evaluate
training and inference times for the smallest and largest benchmark
configurations, providing runtime measurements under the specified
experimental conditions. Overall, the results demonstrate that tree-derived 
structure can serve as an effective inductive bias for neural models while 
preserving an explicit and traceable correspondence between teacher branches 
and neural components, enabling intrinsic interpretability without relying 
on post-hoc explanation methods.
\end{abstract}


\end{frontmatter}

\section{Introduction}

Learning on tabular data has long been dominated by tree-based ensemble methods, particularly gradient boosting~\cite{FRIEDMAN2002367}, with implementations such as XGBoost~\cite{chen2016xgboost} achieving strong and reliable performance across a wide range of tabular benchmarks~\cite{treevsnn}. More recently, neural approaches specifically designed for tabular data have further advanced predictive performance. In particular, TabPFN~\cite{tabpfn} is a transformer-based tabular foundation model that achieves highly competitive results on small to medium-sized tabular classification problems, outperforming established tree-based baselines in its evaluations. These developments demonstrate that tabular prediction can be addressed effectively by different model families. However, predictive performance alone does not determine whether a model's internal reasoning can be represented in a form that is explicit, traceable, and human-understandable.

While individual decision trees are interpretable, strong predictive performance with tree-based methods is typically achieved using ensembles of many trees. In these settings, model behavior arises from the aggregate of many heterogeneous components, which limits the extent to which individual predictions can be traced to a single, compact structure. Neural models can provide flexible and highly optimized representations, but their internal components generally do not correspond explicitly to the decision structures of a tree-based model. Consequently, there is a distinction between achieving high predictive accuracy and providing an explicit, traceable representation of how a model is structured.

This distinction is particularly important in applications where model predictions must be scrutinized and justified to stakeholders, such as domain experts, practitioners, or decision-makers. A common approach to improving the transparency of complex neural models is to apply post-hoc explanation methods after training. Although such methods can provide useful insights into model behavior and support users in understanding AI-assisted decisions, their reliability depends on the method and the context in which it is applied. In particular, model-agnostic interpretation methods can produce misleading conclusions when feature dependencies, interactions, or other relevant properties of the data and model are not adequately considered~\cite{molnar2022}. Inherently interpretable machine learning, instead, seeks to incorporate interpretability into the model design itself, making its structure and behavior accessible to inspection without relying solely on a separate post-hoc explainer~\cite{zschech2026}. This motivates neural architectures that combine effective learning with explicit and inspectable internal structures, providing a complementary route to transparency.

Tree--neural hybrid models offer one approach to this goal, combining the predictive flexibility of neural optimization with the structural interpretability of tree-based models. Rather than competing with purely predictive models on accuracy alone, such approaches aim to retain useful structural information while providing representations that can be inspected at the level of features, decision structures, or model components. Early tree--neural hybrid models, including the ForestNet family~\cite{rguez2019, rguez2020}, have demonstrated that decision-tree structures can be translated into sparse neural architectures with strong interpretability. These models have been successfully applied in several medical domains, including stroke outcome prediction~\cite{rguez2024}, Alzheimer's disease progression~\cite{perez2022}, depression detection in Alzheimer's patients~\cite{perez2023}, and breast cancer ultrasound analysis~\cite{da23-1}. While ForestNet has demonstrated that tree-derived sparsity can provide a meaningful structural prior, its construction and calibration have been closely tied to specific datasets and tasks.

Building on this line of work and earlier tree-to-network mappings~\cite{sethi1990entropynet, rguez2013}, the present paper introduces \textbf{Symbolic Branch Networks (SBNs)}: a principled framework that converts tree ensembles into neural architectures whose hidden units correspond to the branches of the teacher ensemble. The resulting connectivity matrices encode the feature--path and path--class relations induced by the underlying trees. This construction provides an explicit correspondence between the teacher's branch structure and the resulting neural components, enabling feature-level and branch-level interpretation. In the main variant, $\mathrm{SBN}_{1,0}$, the teacher-derived $W_{2}$ is kept fixed while $W_{1}$ is optimized through gradient-based learning, providing a controlled mechanism for adapting the feature-to-branch representation while retaining the teacher-derived branch-to-class parameterization.

In an SBN, every branch (root-to-parent-of-leaf path) of the ensemble becomes a hidden neuron. Two sparse matrices encode the structural information derived from the trees: $W_{1}$ describes which features occur in each branch, while $W_{2}$ describes the association between branches and classes. The resulting network is therefore a direct neural rendering of the ensemble's branch structure. Explicit thresholds and inequality tests are not preserved; instead, the feature-to-branch and branch-to-class relationships are retained through the network's sparsity pattern. This construction provides an explicit correspondence between tree branches and neural components that can be inspected at the feature and branch levels.

Throughout the paper, $\mathrm{SBN}_{a,b}$ denotes whether $W_{1}$ and $W_{2}$, respectively, are trainable, with $a,b \in \{0,1\}$. The notation is summarized in Table~\ref{tab:notation}. The configuration $\mathrm{SBN}_{1,0}$, in which $W_{1}$ is trained while $W_{2}$ remains fixed at its teacher-derived values, is used as the primary SBN configuration.

\begin{table}[ht]
\centering
\caption{SBN parameterization notation. A value of $1$ indicates that the corresponding parameter matrix is trained, whereas $0$ indicates that it remains fixed.}
\label{tab:notation}
\resizebox{\columnwidth}{!}{%
\begin{tabular}{lccc}
\toprule
Configuration & $W_1$ & $W_2$ & Interpretation \\
\midrule
$\mathrm{SBN}_{0,0}$ & Fixed & Fixed & Both matrices fixed \\
$\mathrm{SBN}_{0,1}$ & Fixed & Trained & Only $W_2$ trained \\
$\mathrm{SBN}_{1,0}$ & Trained & Fixed & Primary SBN configuration \\
$\mathrm{SBN}_{1,1}$ & Trained & Trained & Both parameter matrices trained \\
\bottomrule
\end{tabular}
}
\end{table}

Across 29 multiclass datasets from the OpenML CC-18 benchmark, $\mathrm{SBN}_{1,0}$ improves upon the teacher ensemble on 28 of 29 datasets and achieves a mean accuracy of 87.9\%, compared with 84.0\% for the teacher ensemble. A one-sided Wilcoxon signed-rank test across the 29 datasets indicates a statistically significant improvement over the teacher ensemble ($p=4.71\times10^{-7}$). XGBoost achieves a mean accuracy of 87.3\%, while TabPFN achieves 91.3\% on the same benchmark. These comparisons provide predictive reference points for the SBN evaluation, while SBN addresses the additional objective of explicitly incorporating the structure of a tree-based teacher into its neural architecture, providing a traceable correspondence between tree branches and neural components. The $\mathrm{SBN}_{0,0}$ configuration further shows that the teacher-derived structure itself can provide a useful inductive bias, achieving a mean accuracy of 86.1\% while keeping both $W_1$ and $W_2$ fixed.

The main contributions of this work are:
\begin{itemize}
    \item We introduce Symbolic Branch Networks (SBNs), neural architectures
    constructed from the branch structure of decision-tree ensembles,
    providing an explicit correspondence between tree branches and neural
    components.

    \item We introduce $\mathrm{SBN}_{1,0}$ as the primary configuration,
    optimizing the teacher-derived feature-to-branch matrix $W_{1}$ through
    gradient-based learning while keeping the branch-to-class matrix $W_{2}$
    fixed.

    \item We systematically evaluate the four SBN training regimes defined
    by the trainability of $W_{1}$ and $W_{2}$, showing that
    $\mathrm{SBN}_{1,0}$ achieves the highest mean accuracy among the
    considered SBN variants on the multiclass benchmark.

    \item Across 29 multiclass datasets from the OpenML CC-18 benchmark,
    $\mathrm{SBN}_{1,0}$ improves upon the teacher ensemble on 28 of 29
    datasets and achieves a mean accuracy of 87.9\%, compared with 84.0\%
    for the teacher ensemble. XGBoost and TabPFN achieve mean accuracies of
    87.3\% and 91.3\%, respectively, on the same benchmark.

    \item We show that the resulting neural representation retains an
    explicit and traceable correspondence to the teacher-derived branch
    structure, providing a basis for feature-level and branch-level
    interpretation while maintaining competitive predictive performance.

    \item We evaluate local explanation stability across ten independently
    constructed teacher ensembles and characterize training and inference
    times for the smallest and largest evaluated SBN configurations,
    providing additional evidence on explanation consistency and practical
    runtime under the specified experimental conditions.
\end{itemize}

\section{Related Work}

\subsection{Tree-Based Methods for Tabular Data}

Tree-based ensemble methods have long been a dominant approach for learning on tabular data. In particular, gradient-boosted decision trees have consistently demonstrated strong performance across a wide range of real-world benchmarks and applications~\cite{FRIEDMAN2002367,treevsnn}. Their effectiveness has made tree-based ensembles an important reference point for tabular learning, particularly because they can model complex nonlinear relationships while requiring relatively limited feature preprocessing.

Gradient-boosted trees construct ensembles in a stage-wise manner, where each new tree is trained to correct the errors of the existing ensemble. Popular implementations such as XGBoost~\cite{chen2016xgboost} have demonstrated strong predictive performance across diverse tabular datasets. However, although individual decision trees are relatively easy to inspect, the behavior of a large ensemble emerges from the interaction of many heterogeneous trees. This makes it more difficult to trace the behavior of the complete ensemble to a compact, human-understandable structure.

Despite their predictive effectiveness, tree ensembles also have structural properties that distinguish them from neural models. Their decision rules are based on discrete feature thresholds and branching operations, and the models are not naturally differentiable. Consequently, the structure of a trained tree ensemble is not directly amenable to gradient-based optimization or straightforward integration into neural architectures. These properties have motivated research on neural approaches for tabular data, as well as methods that seek to combine tree-based structure with differentiable learning.

\subsection{Neural Networks for Tabular Learning}

Neural networks provide a complementary approach to tabular learning, offering differentiable representations and gradient-based optimization. However, generic neural architectures have historically been less consistently competitive on tabular data than tree-based ensembles~\cite{treevsnn}. This has motivated the development of neural architectures specifically designed for tabular datasets.

More recently, transformer-based approaches have substantially strengthened the performance of neural models on tabular classification. TabPFN~\cite{tabpfn}, for example, is a tabular foundation model trained across large collections of synthetic datasets and has demonstrated strong predictive performance on small- to medium-sized tabular classification problems. These developments demonstrate that neural models can achieve highly competitive predictive performance on tabular data.

Despite these advances, neural architectures generally do not provide an explicit correspondence between their internal components and the decision structures of a tree ensemble. This distinction motivates approaches that seek to combine the optimization flexibility of neural networks with explicit structural information derived from tree-based models.

\subsection{Differentiable Trees and Tree--Neural Hybrids}

A number of approaches have pursued combining decision trees with neural networks in order to leverage the inductive biases of trees while enabling gradient-based optimization. One line of work focuses on differentiable decision trees, where hard split functions are replaced by soft, continuous routing mechanisms that allow end-to-end training via backpropagation~\cite{irsoy2012soft, kontschieder2015deep}. These models preserve a tree-like structure but relax discrete decisions to obtain differentiability.

Related methods extend this idea to ensembles, resulting in neuralized forests or soft decision forests that aggregate the outputs of multiple differentiable trees~\cite{xiao2017nrf, kontschieder2015deep}. While such models enable joint optimization of routing and prediction, they often sacrifice the exact symbolic structure of classical decision trees, and their interpretability can be limited by the use of soft routing and dense parameterization.

\subsection{Tree-to-Network Mappings and Symbolic Neural Models}

Another line of work explores explicit tree--neural hybrids that embed tree
structures into neural architectures without fully relaxing their symbolic
content. Early examples include Entropy Nets~\cite{sethi1990entropynet}, which
represent individual decision trees as neural networks using fixed
connectivity patterns, as well as IG Nets that design partially connected 
neural architectures based on feature relevance, information gain and class
distributions ~\cite{rguez2013}.

The ForestNet family~\cite{rguez2019, rguez2020} extends the idea from individual
trees to ensembles of randomized trees, with tree branches mapped to hidden
neurons and the resulting network topology reflecting the structure of the
teacher ensemble. These methods demonstrate the feasibility of using
tree-derived sparsity as an explicit structural prior and have been applied
to a range of domains, including stroke outcome prediction~\cite{rguez2024},
Alzheimer's disease progression~\cite{perez2022}, depression detection in
Alzheimer's patients~\cite{perez2023}, and breast cancer ultrasound analysis
~\cite{da23-1}.

\subsection{Relation to This Work}

An earlier preprint introduced BranchNet, a prototype for transforming the
branch structure of a tree ensemble into a sparse neural architecture
~\cite{my-arxiv}. The present work builds on this initial formulation. The
basic branch-to-network mapping, in which each tree branch is represented by a
hidden neuron, is also closely related to the construction used in the
ForestNet family~\cite{rguez2019, rguez2020}. Thus, the novelty of the present
work does not lie in introducing the branch-to-neuron mapping itself. Rather,
the present work develops this mapping into a unified and reproducible SBN
framework with an explicit parameterization of the teacher-derived structure,
global calibration parameters, controlled training regimes, and a
standardized experimental protocol.

In the present formulation, the feature-to-branch and branch-to-class
relationships are represented explicitly by the structural matrices $W_1$
and $W_2$, respectively. Global scaling parameters $r_1$ and $r_2$ provide
separate calibration of the two transformations. The training status of the
structural matrices is specified by $\mathrm{SBN}_{a,b}$, where the first
index indicates whether $W_1$ is trainable and the second indicates whether
$W_2$ is trainable. This formulation allows the teacher-derived structure to
remain fixed or be selectively adapted during optimization. The primary
configuration considered in this work is $\mathrm{SBN}_{1,0}$, in which
$W_1$ is optimized while $W_2$ remains fixed at its teacher-derived values.

The ForestNet studies employ different teacher-ensemble and neural-training
configurations across applications, including differences in the number and
depth of trees, teacher model, optimization procedure, learning rate, and
activation function. The present work instead adopts a fixed and
dataset-independent protocol, allowing the proposed $\mathrm{SBN}$ configurations to be
evaluated consistently across a broad collection of multiclass datasets.

The present work also clarifies the meaning of the symbolic structure in $\mathrm{SBN}$.
The mapping does not preserve the original thresholds, inequality tests, or
complete decision-rule semantics of the teacher ensemble. Instead, symbolic
information is represented through the explicit correspondence between
teacher branches and neural components, with the connectivity pattern
encoding feature-to-branch and branch-to-class relationships. The symbolic
aspect, therefore, concerns the explicit structural correspondence inherited
from the teacher ensemble, rather than the preservation of the teacher's complete
decision rules.

\section{Symbolic Branch Networks}
\label{sec:sbn}

Symbolic Branch Networks (SBNs) are neural architectures derived from an
ensemble of decision trees. Instead of learning an unconstrained latent
representation, an SBN fixes the structural support of its hidden layer
according to the symbolic structure extracted from the trees. Each
root-to-parent-of-leaf \emph{branch} becomes a single hidden neuron, while
the numerical values of permitted connections may be fixed or adapted
depending on the $\mathrm{SBN}_{a,b}$ configuration. The two weight matrices
$W_{1}$ and $W_{2}$ encode the feature usage along branches and the class
evidence associated with their corresponding parent-of-leaf nodes.

The resulting architecture provides an explicit structural correspondence
between the teacher-derived branches and the hidden neurons. Each hidden
neuron can therefore be associated with a specific branch extracted from the
teacher ensemble, while its activation is determined by the learned neural
transformation rather than by directly evaluating the original tree
conditions. This correspondence provides a structural basis for inspecting
the model at the branch and feature levels while retaining the flexibility of
gradient-based optimization.

Symbolic Branch Networks can be instantiated with different training regimes depending on whether the weight matrices $W_{1}$ and $W_{2}$ are kept fixed or allowed to adapt during training. We denote these variants by
\[
\mathrm{SBN}_{a,b}, \qquad
\underbrace{a}_{W_{1}}\,,\underbrace{b}_{W_{2}}\in\{0,1\},
\]
where the first index $a$ indicates the training status of $W_{1}$ and the second index $b$ indicates the training status of $W_{2}$. An index of $1$ denotes a trainable matrix, whereas an index of $0$ denotes a fixed matrix. Thus, the four possible variants are
\[
\mathrm{SBN}_{1,1},\quad
\mathrm{SBN}_{1,0},\quad
\mathrm{SBN}_{0,1},\quad
\mathrm{SBN}_{0,0},
\]
corresponding, respectively, to training both matrices, training only $W_{1}$, training only $W_{2}$, and keeping both matrices fixed. In particular, $\mathrm{SBN}_{1,0}$ corresponds to the variant in which $W_{1}$ is adapted during training while $W_{2}$ retains its teacher-derived values, whereas in $\mathrm{SBN}_{0,0}$ both matrices retain their teacher-derived values. The latter should be understood as the variant with fully fixed, teacher-derived parameters rather than as a claim that the network preserves the complete symbolic decision rules of the teacher ensemble.

All four variants are referred to as Symbolic Branch Networks (SBNs), since they share the same branch-based architecture derived from the symbolic structure of the teacher ensemble. The subscripts specify only the training regime of the resulting weight matrices and do not indicate whether the original decision rules of the teacher are preserved.

\subsection{Tree Ensemble Construction}
\label{subsec:treeconstruction}

\begin{figure*}[t]
    \centering
    \begin{subfigure}{0.85\linewidth}
        \centering
        \includegraphics[width=0.85\linewidth]{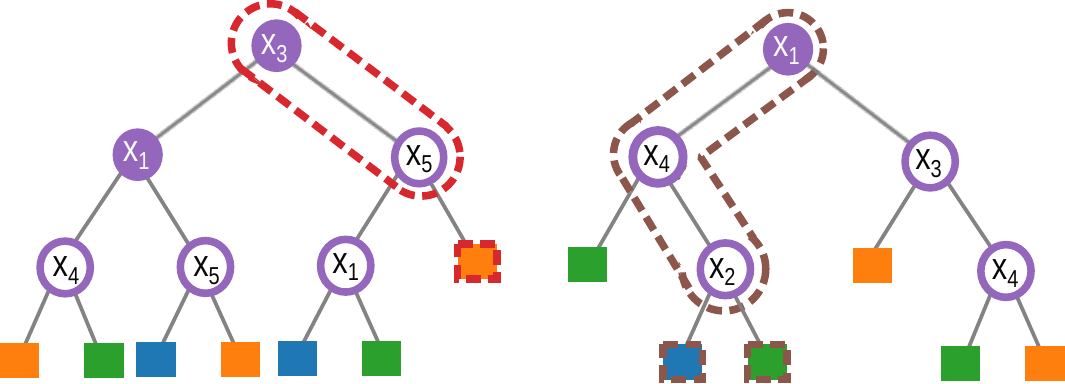}
        \caption{
            Example decision-tree ensemble. Each root–to–parent-of-leaf path is
            extracted as a symbolic branch. The two trees shown contribute four
            branches each, yielding eight branch-neurons in the corresponding
            SBN. One branch per tree is highlighted to illustrate how feature
            usage and class outcomes define the symbolic structure.
        }
        \label{fig:trees}
    \end{subfigure}

    \vspace{1em}

    \begin{subfigure}{0.85\linewidth}
        \centering
        \includegraphics[width=0.85\linewidth]{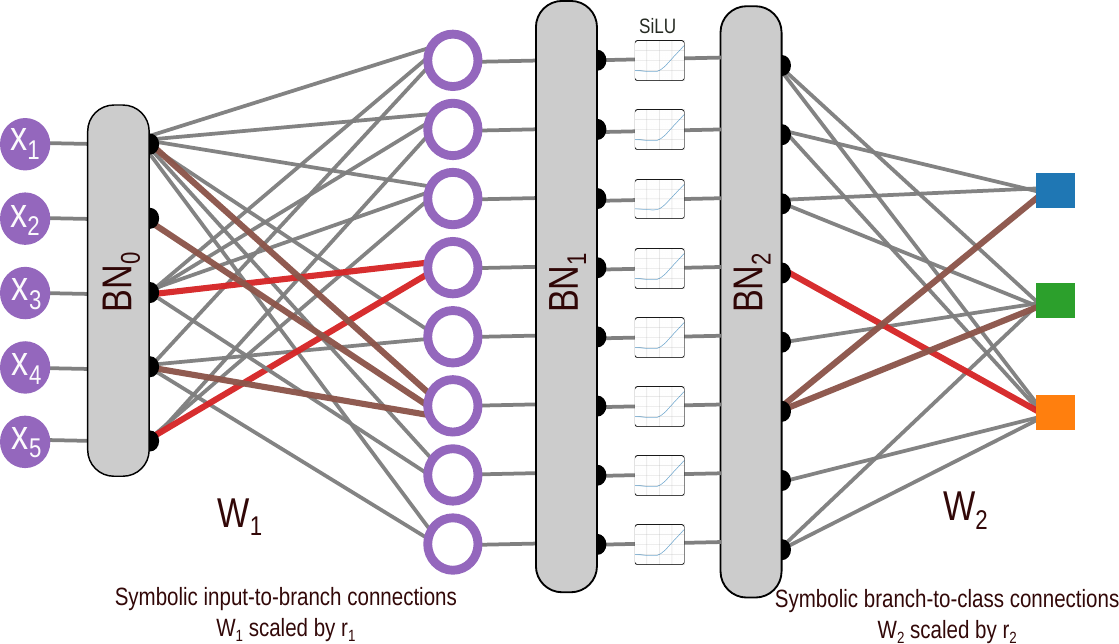}
        \caption{
            The resulting Symbolic Branch Network. Each extracted branch defines
            a sparse input–branch pattern in $W_{1}$ and a sparse
            branch–class pattern in $W_{2}$. The highlighted connections
            correspond exactly to the highlighted tree branches in (a), showing
            how the symbolic structure is inherited by the neural model.
            Batch-normalization layers (BN$_0$, BN$_1$, BN$_2$), the SiLU
            activation, and global scaling factors $r_1$ and $r_2$ provide
            learned calibration while preserving symbolic sparsity.
        }
        \label{fig:network}
    \end{subfigure}

    \caption{
        Construction of a Symbolic Branch Network (SBN) from a decision-tree ensemble.
    }
    \label{fig:overall}
\end{figure*}

To generate the symbolic structure from which SBNs are derived, we train
an ensemble of Extremely Randomized Trees (ExtraTrees) using an incremental
construction protocol described in Algorithm~\ref{alg:ensemble} and illustrated
in Figure~\ref{fig:overall}. We enable \texttt{warm\_start=True} and grow the
ensemble one tree at a time.

Rather than fixing a single tree size, we explicitly control the complexity of
each newly added tree by specifying a maximum number of leaf nodes. For the
$t$-th tree, this limit is set to $2^{d}$, where the exponent $d$ follows a
simple cycling schedule between predefined bounds $d_{\min}$ and $d_{\max}$.
This produces a sequence of trees with varying granularities, ranging from
coarse partitions to more fine-grained symbolic decompositions.

Given $p$ input features and $K$ target classes, we define:
{\footnotesize
\[
d_{\min} = 2, 
\qquad 
d_{\max} = \left\lfloor \log_{2}(p) \right\rfloor + 4,
\qquad 
T_{\max} = K + \left\lfloor \log_{2}(p) \right\rfloor .
\]
}

The ensemble is constructed by incrementally adding $T_{\max}$ trees. At each
step, the current value of $d$ is increased by a fixed amount until it reaches
$d_{\max}$, after which it is reset to $d_{\min}$. This leaf-budget cycling
strategy yields trees with heterogeneous sizes and consequently produces
branch sets of different granularities.

Branches extracted from smaller trees correspond to broader, more general
symbolic rules, while branches from larger trees capture finer subdivisions of
the input space. All root-to-parent-of-leaf paths from the ensemble are then
extracted and converted into hidden neurons, forming the symbolic structure
used by the SBN variants.

\begin{algorithm}[t]
\caption{Tree Ensemble Construction for SBN}
\label{alg:ensemble}
\begin{algorithmic}[1]
\Require training data $(X,y)$ with $p$ input features and $K$ classes
\State $d_{\min}=2,\quad d_{\max}=\lfloor \log_2(p) \rfloor + 4$
\State $T_{\max} = K + \lfloor \log_2(p) \rfloor$
\State Initialize empty ExtraTrees ensemble $E$ with \texttt{warm\_start=True}
\State $d \gets d_{\min}$
\For{$t = 1$ to $T_{\max}$}
    \State $d \gets \min(d + 2, d_{\max})$
    \State Set \texttt{max\_leaf\_nodes} of tree $t$ to $2^d$
    \State Fit tree $t$ and append it to $E$
    \If{$d = d_{\max}$}
        \State $d \gets d_{\min}$
    \EndIf
\EndFor
\State \Return $E$
\end{algorithmic}
\end{algorithm}

\subsection{From Trees to Branch-Neurons}
\label{subsec:tree_to_neurons}

Given an ensemble of decision trees, we extract every unique root-to-parent-of-leaf branch (i.e., paths ending at internal nodes whose children are leaves).
If the ensemble contains $B$ branches in total, then the SBN hidden layer contains exactly $B$ neurons, one per branch. Each branch is characterized by the set of features appearing in its internal nodes and by the class evidence associated with the samples that reach the corresponding leaf-parent node.

\paragraph{Symbolic input-to-branch connectivity matrix $W_{1}$}
For each tree $t$, we first compute a nonnegative feature-weight vector $v^{(t)} \in \mathbb{R}^{d}$, where $v^{(t)}_i$ counts how often feature $i$ appears across all extracted branches of that tree, normalized by the maximum count within the tree. This normalization is performed independently for each tree and does not enforce comparability across trees, reflecting the fact that branches from different trees represent independent symbolic decompositions. 

Let $\mathcal{F}(b_j)$ denote the set of features appearing along branch
$b_j$. For a branch
$b_j$ originating from tree $t$, the corresponding row of $W_1$ is
defined as

{\footnotesize
\[
W_1[j,i] =
\begin{cases}
v^{(t)}_i, & \text{if } i \in \mathcal{F}(b_j),\\
0,         & \text{otherwise}.
\end{cases}
\]
}

This construction yields a highly sparse, nonnegative matrix in which each
branch neuron is connected only to the features appearing along its path, with
weights reflecting the relative structural importance of those features within
the originating tree.

Because all branches from all trees are included, the final $W_{1}$ aggregates
these symbolic feature–branch relationships across the ensemble, yielding a
neural hidden layer whose structure directly mirrors the decomposition
performed by the trees.

\paragraph{Symbolic branch-to-class connectivity matrix $W_{2}$}
For each branch $b_j$ and each class $c$, the entry $W_2[c,j]$ stores a
nonnegative class-evidence value derived from the class counts at the
corresponding tree node, weighted by the fraction of training samples that reach
that node. These values are not normalized and should be interpreted as fixed
branch-level class evidence rather than conditional probabilities. 
This representation retains class-associated evidence from the teacher
ensemble without relying on explicit thresholds or decision rules during
inference.

\paragraph{Symbolic meaning}
Together, the matrices $W_{1}$ and $W_{2}$ define the structural
interpretation of the model:
\begin{itemize}
    \item Rows of $W_{1}$ describe which features are structurally tied to each branch.
    \item Columns of $W_{2}$ describe which classes each branch is associated with.
\end{itemize}
These matrices provide the teacher-derived initialization for the SBN
variants, with the corresponding structural masks preserved during training.

\subsection{Model Architecture}
\label{subsec:architecture}

The SBN architecture implements a calibrated version of the symbolic mappings
defined by $W_{1}$ and $W_{2}$. Let $x \in \mathbb{R}^{d}$ be an input sample. The network
computes:

{\footnotesize	
\[
\begin{aligned}
\mathrm{SBN}(x) =
&\ r_{2} W_{2}\,
  \mathrm{BN}_{2}\!\Big(
    \mathrm{SiLU}\!\big(
      \mathrm{BN}_{1}\!\big(
        r_{1} W_{1}\,\mathrm{BN}_{0}(x)
      \big)
    \big)
  \Big),
\end{aligned}
\]
}

where:
\begin{itemize}
    \item $W_{1}$: symbolic feature-to-branch matrix,
    \item $W_{2}$: symbolic branch-to-class matrix,
    \item $r_{1}, r_{2}$: learned scalar coefficients that calibrate the magnitude of $W_{1}$ and $W_{2}$,
    \item $\mathrm{BN}_{0}, \mathrm{BN}_{1}, \mathrm{BN}_{2}$: batch normalization layers,
    \item $\mathrm{SiLU}$: smooth activation function defined by
$\mathrm{SiLU}(z)=z\,\sigma(z)$, where $\sigma$ is the sigmoid function.
\end{itemize}

The network outputs unnormalized logits. For inference, class probabilities are obtained by applying a Softmax function to the logits, and the predicted class is the argmax of these probabilities.

Every branch neuron corresponds to exactly one symbolic path; the network does not learn new paths nor modify the branching logic. Instead, the model learns to calibrate branch activations through normalization and scaling and, where permitted, to adapt the nonzero weights while preserving the symbolic connectivity structure.

A key property of SBNs is that the symbolic structure inherited from the tree ensemble is preserved by masking during training, which prevents the creation of new connections. Both projection matrices use fixed binary masks:
$m_{1}$ for $W_{1}$ and $m_{2}$ for $W_{2}$.  These masks indicate which feature-to-branch and branch-to-class connections
exist in the underlying trees.  During training, if a matrix is trainable (e.g., $W_{1}$ in $\mathrm{SBN}_{1,0}$), its
updates are always applied element-wise as $W_{1} \leftarrow W_{1} \odot m_{1}$, which prevents the optimizer from creating new connections and forces sparsity. 

In the forward pass, the masked matrices are used whenever the corresponding parameters are trainable:
{\footnotesize
\[
W_{1} \odot m_{1} ~~\text{if } W_{1}\text{ is trainable, else } W_{1};
\]
\[
W_{2} \odot m_{2} ~~\text{if } W_{2}\text{ is trainable, else } W_{2}.
\]
}
Thus, the model can learn calibration through the batch normalization layers without altering the symbolic structure encoded by the tree ensemble.

The scalar coefficients $r_1$ and $r_2$ are learned global calibration parameters that control the overall magnitude of the feature-to-branch and branch-to-class projections, respectively. Both parameters are initialized to $1/\sqrt{d}$, where $d$ is the input dimensionality, and are optimized jointly with the remaining trainable components. This initialization provides a stable starting scale for the symbolic projections while allowing the network to adapt their relative influence during training.

\subsection{Interpretability}
\label{subsec:interpretability}

Because every hidden neuron corresponds to a teacher-derived branch, the SBN
provides an explicit structural basis for both global and local
interpretation. Global interpretation is based on the prevalence of features
across the teacher-derived branches, whereas local interpretation combines
the corresponding branch structure with the learned, sample-specific
contributions of the network.

\subsubsection{Global Interpretability}

Global interpretability in Symbolic Branch Networks is derived directly from
the structural correspondence between the teacher ensemble and the SBN
architecture and does not rely on post-hoc explanation methods. Each hidden
neuron corresponds to a branch extracted from the teacher ensemble. A feature
is therefore considered structurally important at the global level when it
occurs across a large proportion of the extracted branches.

Let $B$ denote the total number of branches and let
$M_1 \in \{0,1\}^{B \times d}$ be the binary sparsity mask associated with the
feature-to-branch matrix $W_1$, where $M_{1,j,i}=1$ if feature $i$ occurs along
branch $j$. We define the structural global importance of feature $i$ as

\[
\mathrm{GI}(i)=\frac{1}{B}\sum_{j=1}^{B}M_{1,j,i}.
\]

Thus, $\mathrm{GI}(i)$ measures the proportion of teacher-derived branches
in which feature $i$ occurs. Features with larger values participate in a
larger fraction of the branch-based representation, whereas features with
smaller values occur in fewer and more specialized branches.

This measure is purely structural. It depends on the teacher-derived
connectivity encoded by $M_1$ and is therefore independent of the numerical
values learned for $W_1$. In particular, when $W_1$ is trainable, optimization
can modify the numerical strength of existing feature-to-branch connections
without changing $\mathrm{GI}(i)$, because the structural mask remains fixed.

The structural global importance therefore provides a global description of
which input features are represented throughout the teacher-derived SBN
architecture. It should be distinguished from sample-specific feature
contributions, which depend on the learned weights, the input values, and the
activation of individual branch-neurons. These local contributions are
considered separately in the local interpretability analysis.

\subsubsection{Local interpretability: explaining a single prediction}
\label{subsec:local_explanations}

Global structural importance describes the teacher-derived architecture
independently of a particular input. To explain an individual prediction,
we instead examine the learned numerical parameters together with the
activation of the branch-neurons for the given sample.

Given a trained SBN and a sample $x\in\mathbb{R}^{d}$, let
\[
z=\mathrm{BN}_0(x),\qquad
h=\phi\!\left(\mathrm{BN}_1(r_1W_1z)\right),\qquad
\tilde{h}=\mathrm{BN}_2(h),
\]
and let the logits be
\[
s=r_2W_2\tilde{h}.
\]

The contribution of branch-neuron $j$ to the predicted class $\hat c$ is
defined as
\[
\alpha_j(x)=\tilde h_j(r_2W_2)_{\hat c,j}.
\]

The branch-neurons with the largest positive values of $\alpha_j(x)$ are selected
for the local explanation, up to the specified number of branches. Because each such neuron corresponds to a
specific teacher-derived branch, the prediction can be traced back to
specific structural components of the SBN.
We focus on positive contributions to identify the branches that support the predicted class, rather than branches that oppose it.

For a selected branch-neuron $j$, the feature-level contribution to its
pre-activation is represented by
\[
a_{j,i}(x)=(r_1W_1)_{j,i}z_i.
\]
This quantity represents the signed feature contribution to the branch neuron's linear projection before the subsequent batch normalization and activation. For the local explanation, the absolute feature contributions are weighted by the corresponding positive branch contributions and aggregated across the selected branches. Features are then ranked by these aggregate scores, with larger scores indicating greater contribution magnitude under this explanation procedure.

The local explanation therefore combines three levels of information:
the teacher-derived branch associated with the neuron, the
sample-specific feature contributions to that neuron, and the neuron's
class-specific contributions at the output layer.

\subsection{End-to-End Pipeline}
\label{subsec:pipeline}

The complete workflow of a Symbolic Branch Network consists of the following
stages:

\begin{enumerate}
    \item \textbf{Tabular preprocessing.}  
    All categorical attributes are converted to numerical or binary form through
    deterministic encoders (e.g., one-hot encoding or ordinal encoding).
    No feature scaling is applied at this stage, since the SBN includes a learned
    input normalization layer (BN$_0$).

    \item \textbf{Tree-ensemble construction.}
    An ExtraTrees ensemble is trained on the processed tabular data.
    A depth-cycling procedure is used to expose the network to heterogeneous
    tree granularities: an internal depth parameter is increased in increments
    of two across sequentially added estimators and reset once a
    dataset-dependent upper limit is exceeded. This parameter determines the theoretical maximum number of leaf nodes as $2^{d}$ for each tree.
    The resulting ensemble therefore contains trees with different structural
    granularities and produces a heterogeneous collection of branch paths.

    \item \textbf{Branch extraction.}  
    Each root-to-parent-of-leaf path is extracted as a symbolic decision branch.
    If the ensemble yields $B$ such branches, the SBN hidden layer has exactly
    $B$ neurons.

    \item \textbf{SBN construction.}  
    The matrices $W_{1}$ (input-to-branch) and $W_{2}$ (branch-to-class) are
    derived directly from symbolic information: feature usage along the branch
    and class evidence at the corresponding parent-of-leaf node.  
    Their binary masks $m_{1}$ and $m_{2}$ encode the resulting sparsity pattern.

    \item \textbf{Prediction and interpretation.}  
    After training, the network outputs calibrated class logits; probabilities are
    obtained by applying a Softmax at inference time. Global feature relevance is 
    obtained from the symbolic structure of $W_{1}$, while sample-level explanations are derived from the branches with the largest positive contributions to the predicted-class logit.

\end{enumerate}

\section{Experiments and results}
\label{sec:experiments}

\subsection{Materials}

We evaluated the proposed method on multiclass classification tasks
from the OpenML-CC18 benchmark suite \cite{oml-benchmarking-suites}. The selection was motivated by the intended scope of the method, which is supervised multiclass classification on tabular data. We therefore considered datasets in which observations are represented as fixed-length feature vectors and classification can reasonably be treated as a tabular learning problem.

Two datasets in OpenML-CC18, CIFAR-10 and Fashion-MNIST, were
excluded from the main experiments. Although both datasets are provided in a tabular representation in OpenML, their features correspond to raw image pixels, and their classification problem involves substantial spatial structure and variation. Exploiting such structure typically requires inductive biases that are not present in general-purpose tabular learning algorithms. In contrast, datasets such as the MFeat collection, which provide fixed-length, engineered image descriptors, were retained because the feature representation itself constitutes the input to the tabular learning problem.

We further restricted the main evaluation to multiclass datasets, consistent with the target setting of the proposed method. Binary classification datasets from CC18 were not included in the primary benchmark because binary classification is outside the specific experimental scope considered here. These datasets and the corresponding results are nevertheless reported in the Appendix to provide a more complete assessment of the method across the CC18 suite. 

Thus, the main benchmark comprises the CC18 datasets that satisfy both criteria: (i) a fixed-length tabular feature representation suitable for conventional tabular learning, and (ii) a multiclass classification target. The resulting evaluation is intended to assess performance in the regime for which the proposed method was designed, rather than to characterize its performance on image or binary-classification problems.

Table~\ref{tab:dataset-summary} provides a high-level overview of the datasets
used in the experiments. Detailed architectural characteristics induced by each
dataset, including model size and sparsity statistics, are reported later in
Section~\ref{subsec:architectural_char}, with corresponding results for binary
classification tasks reported in \ref{app:binary}.

\begin{table*}[t]
\centering
\caption{Summary of datasets used in the experiments.}
\label{tab:dataset-summary}
\begin{tabular}{lcccc}
\toprule
Task type & \# datasets & \# samples (range) & \# features (range) & \# classes \\
\midrule
Multiclass (main study) & 29 & $625$--$70000$ & $4$--$856$ & $3$--$26$ \\
Binary (Appendix) & 30 & $500$--$96320$ & $4$--$1776$ & $2$ \\
\bottomrule
\end{tabular}
\end{table*}

\subsection{Methods}
\label{subsec:methods}
\paragraph{Evaluation protocol}
For each dataset, we perform repeated holdout evaluation over $10$ random seeds. For each seed, the data are split into $70\%$ training, $10\%$ validation, and $20\%$ test sets. The validation split is used for model selection, early stopping, and learning rate scheduling, as applicable to each model. The test split is reserved exclusively for reporting final performance.

All trainable parameters of the proposed models are optimized using AdamW with mild weight decay. Unless stated otherwise, we train for up to $1000$ epochs with a learning rate $\eta = 10^{-1}$, a weight decay $10^{-3}$, and AdamW momentum parameters $(\beta_1,\beta_2)=(0.9,0.98)$. We use a short linear warmup over the first $5$ epochs, during which the learning rate is increased linearly from $0$ to the base learning rate $\eta$. Although the base learning rate is relatively large, training remains stable due to the extreme sparsity of the symbolic matrices, the use of batch normalization at all stages, and the adaptive learning rate reduction provided by the plateau scheduler.

Model selection is performed using a held-out validation set. We apply a ReduceLROnPlateau scheduler on the validation loss with a patience of 35 epochs: if no improvement is observed for 35 consecutive epochs, the learning rate is reduced by a factor of 0.5, with a minimum learning rate of $10^{-4}$. Early stopping is handled independently using a patience of 100 epochs without validation improvement. Model parameters are checkpointed whenever the validation loss reaches a new minimum (improvement margin $10^{-6}$), and the best checkpoint is restored after training.

All experiments were conducted on a system running Ubuntu 24.04.5 LTS, equipped with an Intel Core i5-2400 CPU (4 cores at 3.10 GHz), 31 GiB of RAM, and an NVIDIA GeForce GTX 1080 GPU with 8 GiB of video memory. The software environment comprised Python 3.12.3, PyTorch 2.6.0 with CUDA 12.4 support, scikit-learn 1.9.0, NumPy 2.5.2, Pandas 3.0.5, and XGBoost 3.4.1. CUDA was available to PyTorch. 

\paragraph{Baseline and Models}
The evaluation considers the teacher ensemble used to construct the SBN
architecture together with two external predictive baselines, \textbf{XGBoost}
and \textbf{TabPFN}. The teacher ensemble, denoted \textbf{TeacherSBN}, serves
as the primary reference for evaluating whether the SBN transformation can
improve upon the predictive performance of the model from which its
structure is derived. We evaluate four SBN training configurations,
$\mathrm{SBN}_{0,0}$, $\mathrm{SBN}_{0,1}$, $\mathrm{SBN}_{1,0}$, and
$\mathrm{SBN}_{1,1}$, which differ in whether the teacher-derived matrices
$W_1$ and $W_2$ are kept fixed or optimized during training.

As external reference models, we consider XGBoost, a widely adopted
gradient-boosted tree method for tabular classification~\cite{chen2016xgboost},
and TabPFN, a transformer-based tabular foundation model designed
for tabular prediction~\cite{tabpfn}. These models provide complementary
reference points for assessing predictive performance beyond the teacher
ensemble. The resulting comparisons therefore distinguish between
improvements over the teacher-generated model and differences in predictive
performance relative to established external methods.

For XGBoost, we performed a hyperparameter search over the values listed in Table~\ref{tab:xgboost_hyperparameters}. The search included the default values of all three tuned parameters; all other hyperparameters were kept at their default settings. Early stopping was enabled during training. The configuration was selected based on validation accuracy, and the selected model was evaluated on the held-out test set.

\begin{table}[ht]
\centering
\begin{tabular}{lcc}
\toprule
Parameter & Default value & Search values \\
\midrule
\texttt{n\_estimators} & 100 & 10, 50, 100, 200 \\
\texttt{max\_depth} & 6 & 3, 6, 9 \\
\texttt{learning\_rate} & 0.3 & 0.03, 0.1, 0.3 \\
\bottomrule
\end{tabular}
\caption{XGBoost hyperparameters and the values considered during the search.}
\label{tab:xgboost_hyperparameters}
\end{table}

TabPFN was evaluated using its default configuration, without additional dataset-specific hyperparameter optimization. As a pretrained tabular foundation model, TabPFN differs from conventional machine-learning methods in that its predictions leverage knowledge acquired during pretraining rather than relying solely on dataset-specific parameter optimization. We therefore use the default configuration as a reference for the performance attainable without additional dataset-specific tuning. This protocol provides a practical baseline while avoiding a separate hyperparameter optimization procedure for TabPFN.

\paragraph{Evaluation Metrics}
For multiclass datasets, we report classification accuracy. For binary datasets 
(\ref{app:binary}), we report the area under the ROC curve (AUC). 
Results are computed on the test split for each seed and summarized using
the mean and standard deviation across seeds.

For statistical comparisons, the mean test accuracy across the 10 seeds
is used as the single paired observation for each dataset, yielding
29 paired observations. 
For comparisons between the SBN variants and TeacherSBN, we use one-sided
Wilcoxon signed-rank tests across the 29 datasets, since the primary
hypothesis is directional and concerns whether the SBN configurations
improve upon their teacher. For comparisons with the external methods,
XGBoost and TabPFN, we use two-sided Wilcoxon signed-rank tests because these
comparisons assess whether predictive performance differs without specifying
a direction a priori.

\subsection{Results on Classification Performance}

\subsubsection{SBN against the teacher ensemble}

Table~\ref{tab:accuracy_multiclass} reports the classification accuracy of
the teacher ensemble and the four SBN training configurations across the
29 multiclass datasets. The teacher ensemble achieves a mean accuracy of
84.0\%, while the four SBN variants achieve mean accuracies of 86.1\%,
85.9\%, 87.9\%, and 87.8\% for $\mathrm{SBN}_{0,0}$,
$\mathrm{SBN}_{0,1}$, $\mathrm{SBN}_{1,0}$, and $\mathrm{SBN}_{1,1}$,
respectively. Thus, all four SBN configurations achieve a higher mean
accuracy than their teacher ensemble.

The primary configuration, $\mathrm{SBN}_{1,0}$, obtains the highest mean
accuracy among the four SBN variants, increasing the mean accuracy from
84.0\% for TeacherSBN to 87.9\%. It improves upon the teacher ensemble on
28 of the 29 datasets. The improvement is observed across datasets with
substantially different characteristics, including datasets on which the
teacher already achieves high accuracy. For example, on \textit{car},
$\mathrm{SBN}_{1,0}$ increases accuracy from 93.3\% to 99.9\%, while on
\textit{connect-4} it increases accuracy from 68.9\% to 81.4\%. On
\textit{mnist\_784}, the corresponding increase is from 94.5\% to 97.8\%.
These results indicate that allowing the feature-to-branch matrix $W_1$ to
adapt while keeping the branch-to-class matrix $W_2$ fixed can substantially
improve predictive performance relative to the teacher-derived model.

The results also show that fixing both structural matrices does not prevent
the SBN representation from improving upon the teacher. $\mathrm{SBN}_{0,0}$
achieves a mean accuracy of 86.1\%, compared with 84.0\% for TeacherSBN.
This provides evidence that the branch-derived neural representation can
provide a useful inductive bias even when the structural matrices $W_1$ and
$W_2$ remain fixed and only the remaining trainable calibration parameters
are optimized.
Allowing only $W_2$ to adapt in $\mathrm{SBN}_{0,1}$ results in a mean
accuracy of 85.9\%, whereas allowing only $W_1$ to adapt in
$\mathrm{SBN}_{1,0}$ increases the mean accuracy to 87.9\%. Training both
matrices in $\mathrm{SBN}_{1,1}$ produces a similar mean accuracy of 87.8\%.
Thus, within the configurations considered here, adapting the
feature-to-branch matrix while retaining the teacher-derived
branch-to-class matrix provides the strongest mean predictive performance.

To assess whether the observed improvements over the teacher are consistent
across datasets, we perform one-sided Wilcoxon signed-rank tests on the
paired accuracies of the 29 datasets. The one-sided formulation reflects the
directional hypothesis that the SBN configurations improve upon their
teacher ensemble. The results are reported in
Table~\ref{tab:wilcoxon_teacher}. All four SBN variants show statistically
significant improvements relative to TeacherSBN, with
$p=2.49\times10^{-4}$ for $\mathrm{SBN}_{0,0}$,
$p=3.68\times10^{-4}$ for $\mathrm{SBN}_{0,1}$,
$p=4.71\times10^{-7}$ for $\mathrm{SBN}_{1,0}$, and
$p=8.33\times10^{-7}$ for $\mathrm{SBN}_{1,1}$.

The highest mean accuracy is achieved by $\mathrm{SBN}_{1,0}$, for which the
Wilcoxon statistic is $417$ with $p=4.71\times10^{-7}$. Together with the
improvement on 28 of the 29 datasets and the increase in mean accuracy from
84.0\% to 87.9\%, this supports the use of $\mathrm{SBN}_{1,0}$ as the
primary configuration for the subsequent analyses.

\begin{table*}[ht]
\centering
\caption{Mean accuracy across 10 random seeds for each SBN variant and their teacher
ensemble, TeacherSBN, across 29 multiclass datasets. Values are reported as the mean $\pm$ standard deviation. The
final row reports the mean and standard deviation across datasets.}
\begin{tabular}{lccccc}
\toprule
Dataset & TeacherSBN & $\mathrm{SBN}_{0,0}$ & $\mathrm{SBN}_{0,1}$ & $\mathrm{SBN}_{1,0}$ & $\mathrm{SBN}_{1,1}$ \\
\midrule
analcatdata\_authorship & 0.978 $\pm$ 0.009 & 0.989 $\pm$ 0.009 & 0.991 $\pm$ 0.009 & \textbf{0.993} $\pm$ 0.008 & 0.989 $\pm$ 0.011 \\
balance-scale & 0.840 $\pm$ 0.024 & 0.890 $\pm$ 0.025 & 0.927 $\pm$ 0.042 & \textbf{0.951} $\pm$ 0.020 & 0.947 $\pm$ 0.026 \\
car & 0.933 $\pm$ 0.011 & 0.982 $\pm$ 0.012 & 0.984 $\pm$ 0.008 & \textbf{0.999} $\pm$ 0.002 & \textbf{0.999} $\pm$ 0.002 \\
cmc & 0.514 $\pm$ 0.028 & 0.542 $\pm$ 0.032 & 0.539 $\pm$ 0.037 & \textbf{0.545} $\pm$ 0.043 & 0.538 $\pm$ 0.038 \\
cnae-9 & 0.914 $\pm$ 0.016 & 0.902 $\pm$ 0.019 & 0.912 $\pm$ 0.020 & 0.921 $\pm$ 0.021 & \textbf{0.924} $\pm$ 0.018 \\
connect-4 & 0.689 $\pm$ 0.007 & 0.763 $\pm$ 0.007 & 0.753 $\pm$ 0.008 & 0.814 $\pm$ 0.050 & \textbf{0.818} $\pm$ 0.003 \\
dna & 0.921 $\pm$ 0.007 & 0.943 $\pm$ 0.010 & 0.941 $\pm$ 0.009 & \textbf{0.952} $\pm$ 0.006 & 0.949 $\pm$ 0.011 \\
eucalyptus & 0.596 $\pm$ 0.037 & 0.649 $\pm$ 0.071 & 0.635 $\pm$ 0.048 & 0.599 $\pm$ 0.069 & \textbf{0.613} $\pm$ 0.042 \\
first-order-theorem-proving & \textbf{0.590} $\pm$ 0.012 & 0.535 $\pm$ 0.014 & 0.543 $\pm$ 0.011 & 0.549 $\pm$ 0.019 & 0.546 $\pm$ 0.015 \\
GesturePhaseSegmentation & 0.511 $\pm$ 0.010 & 0.506 $\pm$ 0.061 & 0.542 $\pm$ 0.009 & 0.573 $\pm$ 0.022 & \textbf{0.576} $\pm$ 0.008 \\
jungle\_chess\_2pcs\_raw & 0.763 $\pm$ 0.005 & 0.767 $\pm$ 0.014 & 0.774 $\pm$ 0.012 & \textbf{0.865} $\pm$ 0.004 & 0.851 $\pm$ 0.009 \\
letter & 0.892 $\pm$ 0.003 & 0.937 $\pm$ 0.007 & 0.741 $\pm$ 0.311 & \textbf{0.977} $\pm$ 0.002 & 0.974 $\pm$ 0.003 \\
mfeat-factors & 0.954 $\pm$ 0.013 & 0.964 $\pm$ 0.014 & 0.964 $\pm$ 0.014 & \textbf{0.973} $\pm$ 0.012 & \textbf{0.973} $\pm$ 0.007 \\
mfeat-karhunen & 0.922 $\pm$ 0.015 & 0.951 $\pm$ 0.009 & 0.956 $\pm$ 0.011 & \textbf{0.967} $\pm$ 0.008 & 0.963 $\pm$ 0.009 \\
mfeat-morphological & 0.720 $\pm$ 0.024 & 0.746 $\pm$ 0.023 & 0.744 $\pm$ 0.024 & \textbf{0.745} $\pm$ 0.018 & 0.739 $\pm$ 0.022 \\
mfeat-pixel & 0.963 $\pm$ 0.011 & 0.948 $\pm$ 0.009 & 0.957 $\pm$ 0.011 & 0.964 $\pm$ 0.014 & \textbf{0.969} $\pm$ 0.007 \\
mfeat-zernike & 0.756 $\pm$ 0.019 & 0.812 $\pm$ 0.017 & 0.807 $\pm$ 0.016 & 0.810 $\pm$ 0.016 & \textbf{0.815} $\pm$ 0.015 \\
MiceProtein & 0.968 $\pm$ 0.011 & 0.987 $\pm$ 0.007 & 0.989 $\pm$ 0.009 & 0.994 $\pm$ 0.006 & \textbf{0.995} $\pm$ 0.004 \\
mnist\_784 & 0.945 $\pm$ 0.003 & 0.966 $\pm$ 0.002 & 0.969 $\pm$ 0.002 & \textbf{0.978} $\pm$ 0.001 & \textbf{0.978} $\pm$ 0.002 \\
optdigits & 0.968 $\pm$ 0.005 & 0.975 $\pm$ 0.004 & 0.980 $\pm$ 0.005 & \textbf{0.986} $\pm$ 0.004 & \textbf{0.986} $\pm$ 0.003 \\
pendigits & 0.973 $\pm$ 0.003 & 0.991 $\pm$ 0.002 & 0.992 $\pm$ 0.003 & \textbf{0.995} $\pm$ 0.002 & 0.994 $\pm$ 0.001 \\
segment & 0.928 $\pm$ 0.009 & 0.924 $\pm$ 0.009 & 0.926 $\pm$ 0.008 & 0.929 $\pm$ 0.013 & \textbf{0.931} $\pm$ 0.008 \\
semeion & 0.903 $\pm$ 0.015 & 0.882 $\pm$ 0.013 & 0.891 $\pm$ 0.020 & \textbf{0.922} $\pm$ 0.009 & 0.914 $\pm$ 0.011 \\
splice & 0.942 $\pm$ 0.007 & 0.950 $\pm$ 0.007 & 0.946 $\pm$ 0.008 & 0.958 $\pm$ 0.004 & 0.958 $\pm$ 0.007 \\
steel-plates-fault & 0.728 $\pm$ 0.024 & 0.743 $\pm$ 0.016 & 0.749 $\pm$ 0.016 & 0.750 $\pm$ 0.022 & \textbf{0.753} $\pm$ 0.015 \\
texture & 0.955 $\pm$ 0.006 & 0.992 $\pm$ 0.011 & 0.996 $\pm$ 0.002 & \textbf{0.999} $\pm$ 0.001 & 0.998 $\pm$ 0.001 \\
vehicle & 0.732 $\pm$ 0.030 & 0.837 $\pm$ 0.032 & 0.838 $\pm$ 0.033 & 0.841 $\pm$ 0.021 & \textbf{0.850} $\pm$ 0.020 \\
vowel & 0.937 $\pm$ 0.018 & 0.964 $\pm$ 0.019 & 0.970 $\pm$ 0.015 & \textbf{0.979} $\pm$ 0.012 & \textbf{0.979} $\pm$ 0.014 \\
wall-robot-navigation & 0.930 $\pm$ 0.010 & 0.942 $\pm$ 0.010 & 0.943 $\pm$ 0.024 & \textbf{0.949} $\pm$ 0.006 & 0.940 $\pm$ 0.019 \\
\bottomrule
Average across datasets & 0.840 $\pm$ 0.144 & 0.861 $\pm$ 0.144 & 0.859 $\pm$ 0.144 & 0.879 $\pm$ 0.144 & 0.878 $\pm$ 0.143 \\
\bottomrule
\end{tabular}
\label{tab:accuracy_multiclass}
\end{table*}

\begin{table*}[ht]
\centering
\caption{One-sided Wilcoxon signed-rank tests comparing the
mean accuracy of each SBN variant
against their teacher ensemble,
TeacherSBN, across the 29 multiclass datasets.}
\label{tab:wilcoxon_teacher}
\begin{tabular}{lcc}
\toprule
Comparison & Wilcoxon statistic & $p$-value \\
\midrule
$\mathrm{SBN}_{0,0}$ vs. TeacherSBN & 371 & $2.4938\times10^{-4}$ \\
$\mathrm{SBN}_{0,1}$ vs. TeacherSBN & 367 & $3.6796\times10^{-4}$ \\
$\mathrm{SBN}_{1,0}$ vs. TeacherSBN & 417 & $4.7125\times10^{-7}$ \\
$\mathrm{SBN}_{1,1}$ vs. TeacherSBN & 414 & $8.3260\times10^{-7}$ \\
\bottomrule
\end{tabular}
\end{table*}

\subsubsection{SBN against external baselines}

Table~\ref{tab:accuracy_multiclass_others} compares the primary
$\mathrm{SBN}_{1,0}$ configuration with XGBoost and TabPFN across the same
29 multiclass datasets. $\mathrm{SBN}_{1,0}$ achieves a mean accuracy of
87.9\%, compared with 87.3\% for XGBoost and 91.3\% for TabPFN.
Thus, $\mathrm{SBN}_{1,0}$ achieves a mean accuracy close to that of
XGBoost, whereas TabPFN achieves a higher mean accuracy on this benchmark.

The comparison with XGBoost shows a small descriptive advantage for
$\mathrm{SBN}_{1,0}$. $\mathrm{SBN}_{1,0}$ achieves higher accuracy than
XGBoost on 20 of the 29 datasets. For example,
$\mathrm{SBN}_{1,0}$ substantially exceeds XGBoost on
\textit{balance\-scale} (95.1\% versus 85.7\%),
\textit{vehicle} (84.1\% versus 76.0\%), and
\textit{vowel} (97.9\% versus 90.9\%). XGBoost performs better on other
datasets, including \textit{GesturePhaseSegmentation} (66.7\% versus
57.3\%) and \textit{wall\-robot\-navigation} (99.5\% versus 94.9\%).
Overall, the mean accuracy difference is small, with
$\mathrm{SBN}_{1,0}$ achieving 87.9\% compared with 87.3\% for XGBoost.

TabPFN achieves higher accuracy than $\mathrm{SBN}_{1,0}$ on 28 of the
29 datasets. The only dataset on which $\mathrm{SBN}_{1,0}$ exceeds TabPFN
is \textit{car}, with accuracies of 99.9\% and 99.8\%, respectively.
However, the magnitude of the difference varies considerably across
datasets. For example, TabPFN reaches 79.9\% on
\textit{GesturePhaseSegmentation}, compared with 57.3\% for
$\mathrm{SBN}_{1,0}$, and 95.5\% versus 86.5\% on
\textit{jungle\_chess\_2pcs\_raw}. In contrast, the difference is only
1.0 percentage point on \textit{mnist\_784} (98.8\% versus 97.8\%) and
0.1 percentage point on \textit{texture} (100.0\% versus 99.9\%).
These results show that TabPFN achieves higher predictive performance on
this benchmark, although the magnitude of its advantage over
$\mathrm{SBN}_{1,0}$ is dataset-dependent.

To assess whether the paired differences are consistent across datasets,
we perform two-sided Wilcoxon signed-rank tests between
$\mathrm{SBN}_{1,0}$ and each external baseline. As these comparisons do
not specify a direction of difference a priori, the tests are two-sided.
The results are reported in Table~\ref{tab:wilcoxon_external}. The
comparison between $\mathrm{SBN}_{1,0}$ and TabPFN yields a Wilcoxon
statistic of $1$ and a $p$-value of $7.4505\times10^{-9}$, indicating a
statistically significant paired difference across the 29 datasets. In
contrast, the comparison between $\mathrm{SBN}_{1,0}$ and XGBoost yields a
Wilcoxon statistic of $156$ and a $p$-value of $0.19013$, which does not
indicate a statistically significant paired difference at the conventional
$0.05$ level.

Taken together, the external-baseline results place
$\mathrm{SBN}_{1,0}$ in a similar overall predictive\-performance range to
XGBoost, with a slightly higher mean accuracy but no statistically
significant paired difference across the 29 datasets. TabPFN achieves a
higher mean accuracy and a statistically significant paired difference
relative to $\mathrm{SBN}_{1,0}$, although the magnitude of the difference
varies considerably across datasets. 

\begin{table*}
\centering
\caption{Mean accuracy across 10 random seeds for
SBN$_{1,0}$, TabPFN, and XGBoost on 29 multiclass datasets. Values are reported
as the mean $\pm$ standard deviation. The final row reports the mean and standard
deviation across datasets.}
\begin{tabular}{lccc}
\toprule
Dataset & TabPFN & XGBoost & $\mathrm{SBN}_{1,0}$ \\
\midrule
analcatdata\_authorship & \textbf{1.000} $\pm$ 0.000 & 0.983 $\pm$ 0.007 & 0.993 $\pm$ 0.008 \\
balance-scale & \textbf{0.984} $\pm$ 0.009 & 0.857 $\pm$ 0.030 & 0.951 $\pm$ 0.020 \\
car & 0.998 $\pm$ 0.003 & 0.988 $\pm$ 0.007 & \textbf{0.999} $\pm$ 0.002 \\
cmc & \textbf{0.561} $\pm$ 0.026 & 0.529 $\pm$ 0.034 & 0.545 $\pm$ 0.043 \\
cnae-9 & \textbf{0.937} $\pm$ 0.019 & 0.897 $\pm$ 0.022 & 0.921 $\pm$ 0.021 \\
connect-4 & \textbf{0.850} $\pm$ 0.002 & 0.844 $\pm$ 0.013 & 0.814 $\pm$ 0.050 \\
dna & \textbf{0.965} $\pm$ 0.008 & 0.958 $\pm$ 0.008 & 0.952 $\pm$ 0.006 \\
eucalyptus & \textbf{0.714} $\pm$ 0.029 & 0.656 $\pm$ 0.030 & 0.599 $\pm$ 0.069 \\
first-order-theorem-proving & \textbf{0.615} $\pm$ 0.009 & 0.606 $\pm$ 0.011 & 0.549 $\pm$ 0.019 \\
GesturePhaseSegmentation & \textbf{0.799} $\pm$ 0.006 & 0.667 $\pm$ 0.009 & 0.573 $\pm$ 0.022 \\
jungle\_chess\_2pcs\_raw & \textbf{0.955} $\pm$ 0.003 & 0.869 $\pm$ 0.002 & 0.865 $\pm$ 0.004 \\
letter & \textbf{0.989} $\pm$ 0.002 & 0.961 $\pm$ 0.003 & 0.977 $\pm$ 0.002 \\
mfeat-factors & \textbf{0.988} $\pm$ 0.003 & 0.960 $\pm$ 0.011 & 0.973 $\pm$ 0.012 \\
mfeat-karhunen & \textbf{0.984} $\pm$ 0.006 & 0.946 $\pm$ 0.013 & 0.967 $\pm$ 0.008 \\
mfeat-morphological & \textbf{0.758} $\pm$ 0.022 & 0.713 $\pm$ 0.031 & 0.745 $\pm$ 0.018 \\
mfeat-pixel & \textbf{0.985} $\pm$ 0.008 & 0.957 $\pm$ 0.012 & 0.964 $\pm$ 0.014 \\
mfeat-zernike & \textbf{0.839} $\pm$ 0.018 & 0.780 $\pm$ 0.019 & 0.810 $\pm$ 0.016 \\
MiceProtein & \textbf{1.000} $\pm$ 0.000 & 0.968 $\pm$ 0.008 & 0.994 $\pm$ 0.006 \\
mnist\_784 & \textbf{0.988} $\pm$ 0.001 & 0.977 $\pm$ 0.001 & 0.978 $\pm$ 0.001 \\
optdigits & \textbf{0.995} $\pm$ 0.002 & 0.977 $\pm$ 0.005 & 0.986 $\pm$ 0.004 \\
pendigits & \textbf{0.998} $\pm$ 0.001 & 0.989 $\pm$ 0.002 & 0.995 $\pm$ 0.002 \\
segment & \textbf{0.949} $\pm$ 0.005 & 0.929 $\pm$ 0.011 & 0.929 $\pm$ 0.013 \\
semeion & \textbf{0.956} $\pm$ 0.004 & 0.916 $\pm$ 0.015 & 0.922 $\pm$ 0.009 \\
splice & \textbf{0.971} $\pm$ 0.005 & 0.962 $\pm$ 0.007 & 0.958 $\pm$ 0.004 \\
steel-plates-fault & \textbf{0.833} $\pm$ 0.014 & 0.783 $\pm$ 0.018 & 0.750 $\pm$ 0.022 \\
texture & \textbf{1.000} $\pm$ 0.000 & 0.981 $\pm$ 0.004 & 0.999 $\pm$ 0.001 \\
vehicle & \textbf{0.884} $\pm$ 0.028 & 0.760 $\pm$ 0.036 & 0.841 $\pm$ 0.021 \\
vowel & \textbf{0.996} $\pm$ 0.004 & 0.909 $\pm$ 0.028 & 0.979 $\pm$ 0.012 \\
wall-robot-navigation & 0.986 $\pm$ 0.003 & \textbf{0.995} $\pm$ 0.002 & 0.949 $\pm$ 0.006 \\
\bottomrule
Average across datasets & \textbf{0.913} $\pm$ 0.118 &0.873 $\pm$ 0.129 & 0.879 $\pm$ 0.144 \\

\bottomrule
\end{tabular}
\label{tab:accuracy_multiclass_others}
\end{table*}

\begin{table*}[ht]
\centering
\caption{Two-sided Wilcoxon signed-rank tests comparing the mean accuracy of
$\mathrm{SBN}_{1,0}$ against TabPFN and XGBoost across the 29
multiclass datasets.}
\label{tab:wilcoxon_external}
\begin{tabular}{lcc}
\toprule
Comparison & Wilcoxon statistic & $p$-value \\
\midrule
$\mathrm{SBN}_{1,0}$ vs. TabPFN & 1 & $7.4505\times10^{-9}$ \\
$\mathrm{SBN}_{1,0}$ vs. XGBoost & 156 & $1.9013\times10^{-1}$ \\
\bottomrule
\end{tabular}
\end{table*}

\subsection{Results on Global Interpretability}

The global interpretability analysis examines the structural feature
importance induced by the teacher-derived branch connectivity. For each
feature, the structural importance is computed from the binary mask $M_1$ as
the proportion of extracted branch-neurons in which that feature occurs.
Because $M_1$ is fixed during training, this measure describes the structural
representation inherited from the teacher ensemble rather than the learned
numerical strength of individual connections.

\begin{figure*}[!ht]
    \centering

        \includegraphics[width=0.8\linewidth]{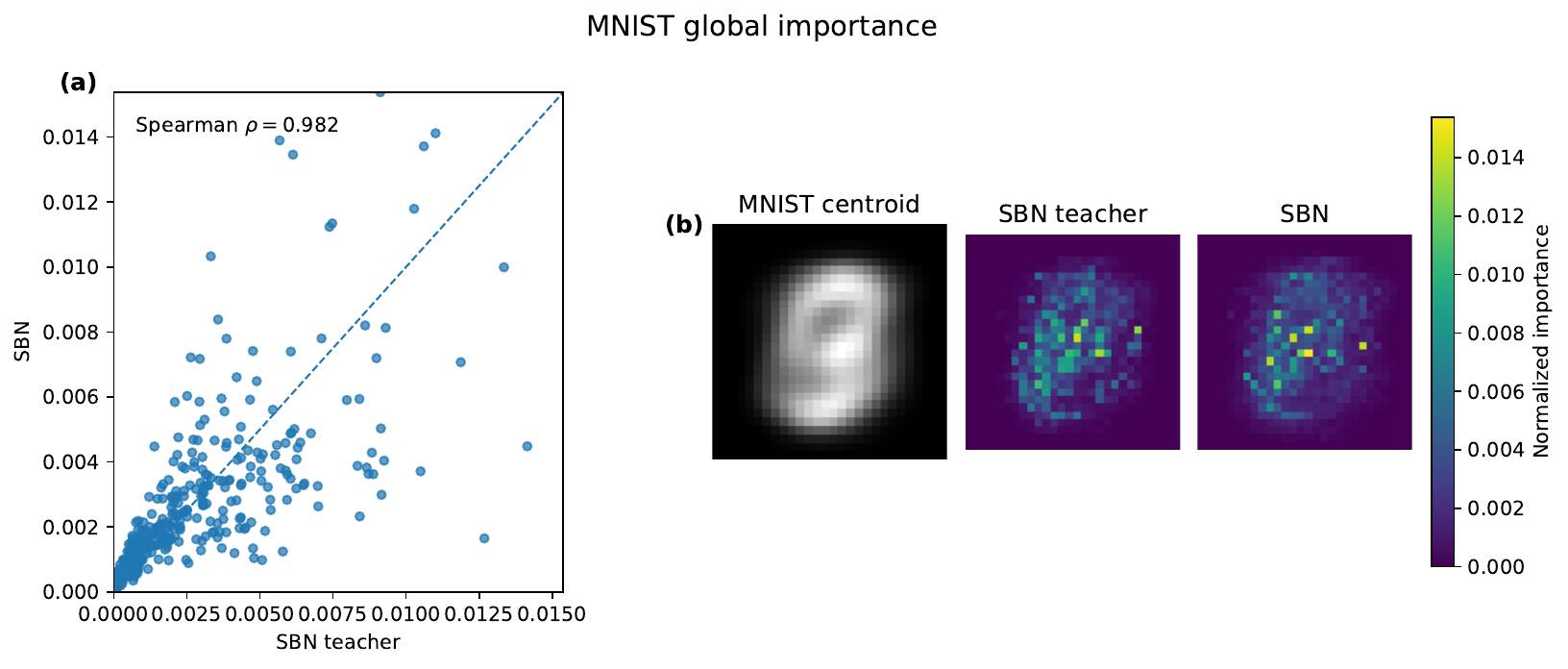}

    \caption{Global structural feature importance on MNIST. (a) Comparison
between teacher-ensemble feature importance and SBN structural feature
importance for individual input features. The two importance profiles exhibit
a Spearman rank correlation of $\rho=0.982$. (b) Spatial visualization of the
MNIST centroid and the corresponding normalized feature-importance maps for
the teacher ensemble and the SBN structural representation.}
\label{fig:global}
\end{figure*}

Figure~\ref{fig:global}(a) compares the structural feature importance of
SBN with the feature importance reported by the teacher ensemble on
\textit{mnist\_784}. The two measures exhibit a strong positive rank
correlation, with a Spearman correlation of $\rho=0.982$. Thus, features
that occur frequently across the teacher-derived SBN branches tend to
coincide with features receiving high feature importance from the teacher
ensemble.

The two measures nevertheless capture different properties. The teacher
ensemble importance is derived from the predictive tree model, whereas SBN
structural importance measures the prevalence of each feature across the
branch-based architecture.
The observed agreement indicates that features occurring frequently in the
teacher-derived branch structure tend to coincide with features receiving
high importance in the teacher ensemble.

Figure~\ref{fig:global}(b) provides a spatial representation of the same
comparison. The MNIST centroid is shown together with the teacher-derived
feature importance and the SBN structural feature importance. The two
importance maps exhibit similar spatial patterns, with greater structural
importance concentrated in the central regions of the digit representation.

Because the SBN structural importance is determined by $M_1$, optimization
of $W_1$ does not change this global structural measure. Instead, training
can modify the numerical strength of the existing feature-to-branch
connections. Such sample-dependent effects are examined separately in the
local interpretability analysis.

\subsection{Results on Local Interpretability}
\label{subsec:local_interpretability}

To investigate whether the branch structure of an SBN provides a useful basis for explaining individual predictions, we examine a single prediction from the Vehicle Silhouettes dataset. The dataset contains four vehicle classes (bus, Opel, Saab, and van), represented by geometric descriptors of the vehicle silhouette. The features used in this analysis are summarized in Table~\ref{tab:vehicle_features}. We select a correctly classified Saab instance and trace the highest-ranked branch-neurons contributing to its prediction.

\begin{table*}[!ht]
\centering
\caption{Geometric features used for vehicle classification.}
\label{tab:vehicle_features}
\small
\begin{tabular}{p{0.34\linewidth} p{0.58\linewidth}}
\toprule
\textbf{Feature} & \textbf{Interpretation} \\
\midrule
\texttt{COMPACTNESS} &
Overall compactness of the vehicle silhouette. \\

\texttt{CIRCULARITY} &
Degree to which the silhouette approaches a circular shape. \\

\texttt{DISTANCE\_CIRCULARITY} &
Circularity measured from distances between the centroid and boundary. \\

\texttt{RADIUS\_RATIO} &
Ratio between characteristic radii of the silhouette. \\

\texttt{PR.AXIS\_ASPECT\_RATIO} &
Aspect ratio defined with respect to the principal axis. \\

\texttt{MAX.LENGTH\_ASPECT\_RATIO} &
Aspect ratio derived from the maximum silhouette length. \\

\texttt{SCATTER\_RATIO} &
Relative spread of the silhouette around its principal axes. \\

\texttt{ELONGATEDNESS} &
Degree to which the silhouette is elongated. \\

\texttt{PR.AXIS\_RECTANGULARITY} &
Degree of rectangularity relative to the principal axis. \\

\texttt{MAX.LENGTH\_RECTANGULARITY} &
Degree of rectangularity based on maximum silhouette dimensions. \\

\texttt{SCALED\_VARIANCE\_MAJOR} &
Variation of the silhouette along the major axis, after scaling. \\

\texttt{SCALED\_VARIANCE\_MINOR} &
Variation of the silhouette along the minor axis, after scaling. \\

\texttt{SCALED\_RADIUS\_OF\_GYRATION} &
Spatial spread of the silhouette around its center, after scaling. \\

\texttt{SKEWNESS\_ABOUT\_MAJOR} &
Asymmetry of the silhouette along the major axis. \\

\texttt{SKEWNESS\_ABOUT\_MINOR} &
Asymmetry of the silhouette along the minor axis. \\

\texttt{KURTOSIS\_ABOUT\_MAJOR} &
Concentration of the silhouette distribution along the major axis. \\

\texttt{KURTOSIS\_ABOUT\_MINOR} &
Concentration of the silhouette distribution along the minor axis. \\

\texttt{HOLLOWS\_RATIO} &
Amount of empty or concave area within the silhouette. \\
\bottomrule
\end{tabular}
\end{table*}

\begin{figure*}[!ht]
    \centering
        \includegraphics[width=0.84\linewidth]{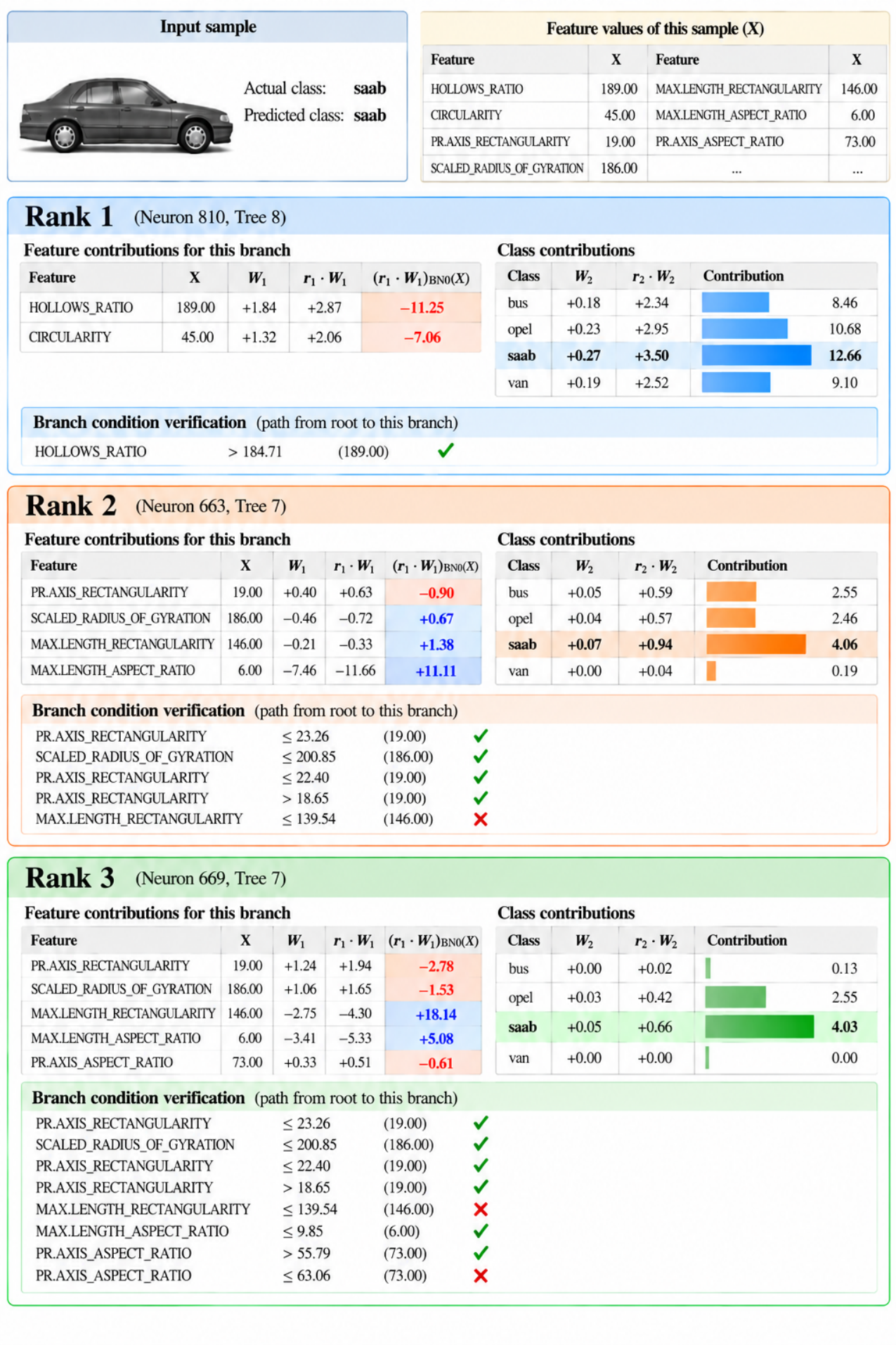}
    \caption{Local interpretation of a correctly classified Saab instance from the Vehicle Silhouettes dataset. The three highest-ranked branch-neurons are shown together with their associated features, branch conditions, and class-specific contributions.}
\label{fig:local}
\end{figure*}

For the analyzed instance, the $\mathrm{SBN}_{1,0}$ predicts \textit{Saab}, which is also the ground-truth class. The local explanation identifies the branch-neurons with the largest contributions to the prediction and exposes both their associated input features and their class-specific effects. Figure~\ref{fig:local} shows the resulting local explanation. This provides a direct correspondence between an individual prediction and specific components of the tree-derived neural architecture.

The first-ranked neuron (Neuron 810) is associated primarily with \texttt{HOLLOWS\_RATIO} and \texttt{CIRCULARITY}. In particular, the observed value of \texttt{HOLLOWS\_RATIO} is 189, satisfying the corresponding teacher-tree condition \texttt{HOLLOWS\_RATIO} $> 184.71$. The corresponding neuron contributes positively to all four classes, but its contribution is largest for Saab ($+12.66$), followed by Opel ($+10.68$), van ($+9.10$), and bus ($+8.46$). Thus, this branch provides strong evidence for the vehicle-like representation captured by the neuron, while providing its strongest class-specific contribution to Saab. Importantly, the contribution should not be interpreted as evidence unique to Saab, since the same neuron also contributes positively to the alternative classes.

The second-ranked neuron (Neuron 663) provides a more selective class contribution. Its representation involves \texttt{PR.AXIS\_RECTANGULARITY}, \texttt{SCALED\_RADIUS\_OF\_GYRATION}, \texttt{MAX.LENGTH\_RECTANGULARITY}, and \texttt{MAX.LENGTH\_ASPECT\_RATIO}. For this instance, the neuron contributes $+4.06$ to Saab, compared with $+2.46$ for Opel, $+2.55$ for bus, and $+0.19$ for van. The difference between the Saab and Opel contributions is particularly relevant because these two classes constitute the more fine-grained distinction within the car-like vehicle categories. This branch therefore provides evidence that is more discriminative between the competing classes than the first-ranked neuron.

The third-ranked neuron (Neuron 669) provides another example of this Saab-versus-Opel distinction. Its parent split is associated with a teacher-tree branch leading toward the Saab class, with the condition
\texttt{SCALED\_RADIUS\_OF\_GYRATION} $> 165.47$. The observed value is 186, satisfying this condition. The neuron contributes $+4.03$ to Saab and $+2.55$ to Opel, while its contributions to bus and van are close to zero ($+0.13$ and $+0.00$, respectively). The local explanation therefore identifies \texttt{SCALED\_RADIUS\_OF\_GYRATION} as part of a learned representation that separates the Saab class from the other vehicle categories, particularly from Opel.

Taken together, the three highest-ranked neurons provide complementary evidence. The first neuron produces a relatively broad response across classes, whereas the second and third neurons provide increasingly selective evidence for Saab. Across all three neurons, the contribution to Saab is larger than the contribution to Opel, the closest competing class in this local explanation. This illustrates how the SBN representation can expose not only which features are involved in a prediction, but also how the corresponding branch-neurons affect each candidate class.

The branch conditions should be interpreted as structural information associated with the corresponding tree-derived neuron rather than as a conventional logical decision-tree path. In particular, some conditions reported for a neuron may not be simultaneously satisfied by the input instance. The local explanation therefore combines two complementary forms of information: the branch conditions identify the decision structure from which the neuron originated, while the feature and class contributions quantify how the learned neural representation responds to the specific instance. This distinction is important because $\mathrm{SBN}_{1,0}$ optimizes $W_1$ after the tree-derived architecture has been constructed, and therefore the learned neural representation need not preserve the original tree thresholds as exact logical rules.

Overall, this example demonstrates that the tree-derived architecture provides a traceable route from an individual prediction to a small set of branch-neurons, their associated features, and their class-specific contributions. Rather than relying solely on post-hoc feature attribution, the explanation is grounded in the structural components of the SBN itself. The example consequently illustrates the intended local interpretability of SBNs: an individual prediction can be inspected in terms of the learned representations associated with specific branches of the teacher ensemble and their relative contributions to the predicted and competing classes.

\subsubsection{Prediction and feature stability.}

To evaluate the stability of predictions and local explanations with respect to the random seed used to construct the teacher ensemble, we performed ten runs using the same training data and the same Vehicle instance in Fig.~\ref{fig:local}. The random seed used to build the teacher ensemble was varied across runs, resulting in ten independently constructed teacher ensembles and their corresponding $\mathrm{SBN}_{1,0}$ models. For each run, we recorded the predicted class and compared it with the ground-truth label. We then identified the three branch-neurons with the largest positive contributions to the predicted-class logit and aggregated their weighted feature contributions to obtain a local feature-importance ranking. To summarize explanation stability, we counted the number of runs in which each feature appeared among the top three and top five features. We also report the median and mean feature ranks across runs, using the rank aggregation procedure described above.

All ten models correctly classified the selected instance as Saab. The predicted class was therefore unchanged across the ten initializations, indicating prediction stability for this instance. However, prediction stability does not necessarily imply explanation stability, since different models can produce the same prediction while assigning different levels of importance to individual features.

Table~\ref{tab:vehicle_feature_stability} summarizes the local feature rankings across the ten runs. \texttt{HOLLOWS\_RATIO} is the most consistently selected feature: it appears among the top three features in nine runs and among the top five in all ten. \texttt{MAX.LENGTH\_RECTANGULARITY} is also consistently selected, appearing in the top three in seven runs and in the top five in nine. \texttt{COMPACTNESS} appears in the top three in four runs and in the top five in six. These frequencies indicate that the leading features recur across independently trained models, although their relative ordering is not identical across runs.

\begin{table*}[t]
\centering
\caption{Stability of local feature rankings across ten
$\mathrm{SBN}_{1,0}$ runs for the selected Vehicle example.
The top-$k$ columns report the number of runs in which each
feature appears among the top $k$ features.}
\label{tab:vehicle_feature_stability}
\small
\begin{tabular}{lcccc}
\hline
Feature & Runs in top 3 & Runs in top 5
        & Median rank & Mean rank \\
\hline
\texttt{HOLLOWS\_RATIO} & 9 & 10 & 1 & 1.60 \\
\texttt{MAX.LENGTH\_RECTANGULARITY} & 7 & 9 & 2 & 2.67 \\
\texttt{COMPACTNESS} & 4 & 6 & 2.5 & 2.83 \\
\texttt{CIRCULARITY} & 2 & 4 & 5 & 4.20 \\
\texttt{MAX.LENGTH\_ASPECT\_RATIO} & 2 & 4 & 5 & 6.71 \\
\texttt{KURTOSIS\_ABOUT\_MINOR} & 2 & 3 & 3 & 3.00 \\
\texttt{PR.AXIS\_RECTANGULARITY} & 1 & 3 & 5.5 & 5.17 \\
\texttt{SKEWNESS\_ABOUT\_MAJOR} & 1 & 3 & 4 & 4.00 \\
\texttt{ELONGATEDNESS} & 0 & 3 & 5 & 5.25 \\
\texttt{RADIUS\_RATIO} & 1 & 1 & 7 & 6.40 \\
\texttt{SCALED\_VARIANCE\_MAJOR} & 1 & 1 & 8.5 & 8.00 \\
\texttt{KURTOSIS\_ABOUT\_MAJOR} & 0 & 1 & 7 & 6.75 \\
\texttt{SCALED\_RADIUS\_OF\_GYRATION} & 0 & 1 & 6 & 5.33 \\
\texttt{SCALED\_VARIANCE\_MINOR} & 0 & 1 & 8 & 8.00 \\
\texttt{DISTANCE\_CIRCULARITY} & 0 & 0 & 8 & 7.33 \\
\texttt{PR.AXIS\_ASPECT\_RATIO} & 0 & 0 & 8 & 8.25 \\
\texttt{SKEWNESS\_ABOUT\_MINOR} & 0 & 0 & 8 & 8.50 \\
\texttt{SCATTER\_RATIO} & 0 & 0 & 9 & 8.86 \\
\hline
\end{tabular}
\end{table*}

The remaining features are selected less consistently. For example, \texttt{CIRCULARITY} and \texttt{MAX.LENGTH\_ASPECT\_RATIO} each appear among the top five features in four runs, while several features do not enter the top five in any run. The rank statistics complement these selection frequencies by describing the relative positions of features in the aggregated rankings. Overall, the results indicate partial stability of local feature explanations: the most prominent features are recovered consistently, whereas the composition and ordering of the remaining features vary with model initialization. Because the analysis considers only one selected instance, these results characterize prediction and explanation stability for that instance and should not be interpreted as evidence of stability across the entire Vehicle dataset.

The stability analysis provides complementary evidence regarding the consistency of SBN explanations at the local and global levels. For the selected Vehicle Silhouettes instance, the repeated runs show consistent predictions and recurring high-ranking features, although the exact feature rankings are not identical across runs. This indicates that the principal features contributing to this particular prediction are relatively stable under the evaluated runs. At the global level, the high Spearman correlation between the feature-importance rankings of SBN and its teacher on MNIST indicates strong agreement between the two models' global importance profiles. However, these findings should be interpreted within the scope of the experiments conducted. The local stability analysis considers a single instance, while the global comparison measures agreement with the teacher rather than the correctness of the explanations against an independent ground truth. Therefore, the results provide preliminary evidence of explanation consistency but do not establish general explanation reliability across samples, datasets, or training conditions.

\subsection{Architectural Characteristics}
\label{subsec:architectural_char}

\subsubsection{Multi-Class Architectural Details}

The architectural characteristics of SBN for multi-class classification
tasks are summarized in Table~\ref{tab:sbn_multi}. The table reports, for
each dataset, the number of classes, features, and samples, together with the
range of hidden neurons induced by the extracted branches and the minimum and
maximum sparsity ratios of the symbolic matrices $W_1$ (input-to-branch) and
$W_2$ (branch-to-class).

For instance, the \texttt{mfeat-karhunen} dataset, with 64 features and 2000
samples, yields between 2677 and 2911 hidden neurons, with $W_1$ sparsity ranging
from 86.39\% to 85.32\% and $W_2$ sparsity from 66.95\% to 68.22\%. In contrast, the
\texttt{cmc} dataset, a 3-class problem with only 9 features and 1473 samples,
produces a much more compact architecture with 485 to 504 hidden neurons and
lower sparsity levels (28.59\%–34.16\% for $W_1$ and 13.62\%–17.04\% for $W_2$).

Overall, high sparsity levels are observed in several datasets, particularly those with high input dimensionality, as
exemplified by \texttt{mnist\_784} (up to 98.07\% $W_1$ sparsity) and \texttt{cnae-9}
(up to 95.78\% $W_1$ sparsity). This reflects SBN’s ability to construct
large but structured models whose connectivity is strictly constrained by the
symbolic decomposition of the teacher ensemble. Each hidden neuron corresponds
to a root-to-parent-of-leaf path, ensuring that model size grows with structural
complexity rather than dense parameterization.

Despite substantial variation in architectural scale and sparsity across
datasets, the resulting SBN architectures remain explicitly structured by
the teacher-derived branch connectivity. 
The fixed structural masks preserve the teacher-derived feature-to-branch
and branch-to-class connectivity while allowing the corresponding numerical
parameters and calibration parameters to be optimized when permitted by the
training configuration.

\begin{table*}[htbp]
\centering
\small

\caption{Summary of SBN architectural characteristics for multi-class datasets.}
\begin{tabular}{rrrrcccccc}
\toprule
                    &    No. of   &   No. of        &                    Total    & \multicolumn{2}{c}{Hidden neurons} & \multicolumn{2}{c}{$W_1$ sparsity (\%)} & \multicolumn{2}{c}{$W_2$ sparsity (\%)} \\
Dataset             & classes & feats & samples & min & max & min & max & min & max \\
\midrule
analcatdata\_authorship & 4 & 70 & 841 & 563 & 627 & 88.89 & 89.53 & 38.40 & 39.88 \\
balance-scale & 3 & 4 & 625 & 331 & 362 & 2.26 & 3.75 & 18.17 & 20.04 \\
car & 4 & 21 & 1728 & 781 & 896 & 54.19 & 55.96 & 44.27 & 45.51 \\
cmc & 3 & 9 & 1473 & 485 & 504 & 28.59 & 34.16 & 13.62 & 17.04 \\
cnae-9 & 9 & 856 & 1080 & 3376 & 3556 & 95.41 & 95.78 & 46.24 & 49.11 \\
connect-4 & 3 & 42 & 67557 & 1955 & 1990 & 72.60 & 74.50 & 3.45 & 4.44 \\
dna & 3 & 180 & 3186 & 1571 & 1866 & 93.83 & 94.21 & 25.19 & 27.27 \\
eucalyptus & 5 & 91 & 736 & 1657 & 1749 & 85.97 & 87.28 & 45.61 & 46.52 \\
first-order-theorem-proving & 6 & 51 & 6118 & 5332 & 5460 & 72.22 & 75.23 & 47.03 & 48.20 \\
GesturePhaseSegmentation & 5 & 32 & 9873 & 2151 & 2196 & 60.58 & 64.52 & 26.43 & 29.24 \\
jungle\_chess\_2pcs\_raw & 3 & 6 & 44819 & 471 & 493 & 16.18 & 24.02 & 4.91 & 11.48 \\
letter & 26 & 16 & 20000 & 6771 & 6815 & 46.24 & 49.21 & 72.55 & 73.43 \\
mfeat-factors & 10 & 216 & 2000 & 2186 & 2312 & 95.64 & 95.84 & 71.16 & 71.85 \\
mfeat-karhunen & 10 & 64 & 2000 & 2677 & 2911 & 84.39 & 85.32 & 66.95 & 68.22 \\
mfeat-morphological & 10 & 6 & 2000 & 1018 & 1046 & 12.87 & 24.84 & 64.73 & 68.28 \\
mfeat-pixel & 10 & 240 & 2000 & 1994 & 2138 & 96.49 & 96.59 & 72.01 & 72.57 \\
mfeat-zernike & 10 & 47 & 2000 & 3237 & 3384 & 76.69 & 77.66 & 69.68 & 70.51 \\
MiceProtein & 8 & 77 & 1080 & 1527 & 1649 & 87.96 & 88.47 & 64.00 & 65.88 \\
mnist\_784 & 10 & 784 & 70000 & 47229 & 47799 & 98.04 & 98.07 & 69.18 & 69.42 \\
optdigits & 10 & 64 & 5620 & 3930 & 4183 & 84.15 & 84.54 & 70.28 & 71.28 \\
pendigits & 10 & 16 & 10992 & 2697 & 2789 & 46.87 & 48.59 & 71.30 & 72.23 \\
segment & 7 & 16 & 2310 & 1361 & 1462 & 47.87 & 51.90 & 65.15 & 66.58 \\
semeion & 10 & 256 & 1593 & 2557 & 2713 & 96.33 & 96.43 & 70.79 & 71.29 \\
splice & 3 & 284 & 3190 & 2182 & 2413 & 95.67 & 96.28 & 26.09 & 27.78 \\
steel-plates-fault & 7 & 27 & 1941 & 1827 & 1908 & 63.81 & 66.57 & 61.17 & 62.54 \\
texture & 11 & 40 & 5500 & 2529 & 2649 & 74.39 & 75.58 & 72.66 & 73.87 \\
vehicle & 4 & 18 & 846 & 804 & 866 & 54.87 & 57.36 & 38.84 & 41.59 \\
vowel & 11 & 27 & 990 & 1810 & 1893 & 66.85 & 69.94 & 70.62 & 72.18 \\
wall-robot-navigation & 4 & 24 & 5456 & 1400 & 1660 & 60.94 & 64.26 & 40.52 & 42.90 \\
\bottomrule
\end{tabular}
\label{tab:sbn_multi}
\end{table*}

Architectural characteristics for binary classification tasks are reported in \ref{app:binary}, Table~\ref{tab:sbn_binary}.

\subsubsection{Training and Inference Times}

We evaluated the runtimes of the smallest and largest SBN configurations in our benchmark, corresponding to balance-scale and MNIST, respectively. Each configuration was measured over ten runs. The balance-scale dataset contains 625 instances across the evaluated splits, comprising 437 training instances, 63 validation instances, and 125 test instances, with 4 input features. MNIST contains 70,000 instances across the evaluated splits, comprising 49,000 training instances, 7,000 validation instances, and 14,000 test instances, with 784 input features.

As described in Section~\ref{subsec:tree_to_neurons}, the number of trees is determined by the number of classes and input features, with
$T_{\max}=K+\lfloor\log_2(p)\rfloor$. The maximum number of leaf nodes per tree is also bounded by the feature-dependent depth schedule, since
$d\leq\lfloor\log_2(p)\rfloor+4$ and each tree is constrained by
$\texttt{max\_leaf\_nodes}=2^d$. Consequently, for fixed $p$ and $K$, the number of trees and the maximum structural size of the ensemble are bounded independently of the number of training instances. Since the number of branch neurons depends on the branches extracted from the ensemble, the resulting SBN architecture is likewise structurally bounded by the feature dimensionality and number of classes under these construction rules, rather than growing directly with the training-set size. This structural bound does not imply that training cost is independent of the number of samples, as fitting the tree ensemble and optimizing the student still require processing the training data.

As shown in Table~\ref{tab:sbn_training_time}, the mean total training pipeline time was $7.585 \pm 1.264$~s for balance-scale and $1698.733 \pm 286.043$~s (approximately 28.3 min) for MNIST. The measured pipeline comprised teacher ensemble training, SBN architecture construction, and student training. Student training was the dominant computational cost, accounting for approximately 97.5\% of the total time for balance-scale and 96.8\% for MNIST. These timings include validation evaluations performed during student optimization within the timed training call. The observed variability in training time may partly reflect differences in the architectures generated by the teacher ensemble across runs, including variation in the number of hidden neurons.

\begin{table*}[t]
\centering
\caption{Training and class-prediction inference times
(mean $\pm$ standard deviation over 10 runs) for the smallest and
largest evaluated SBN configurations. Balance-scale has 4 input
features and 437 training, 63 validation, and 125 test instances.
MNIST has 784 input features and 49,000 training, 7,000 validation,
and 14,000 test instances. Classification inference measures class
label prediction on the complete test set and excludes probability
estimation.}
\label{tab:sbn_training_time}
\begin{tabular}{lcc}
\hline
\textbf{Component} & \textbf{Balance-scale} & \textbf{MNIST} \\
\hline
\textit{Training time (s)} & &  \\
Teacher training
    & $0.028 \pm 0.007$
    & $7.823 \pm 0.390$ \\
Architecture construction
    & $0.163 \pm 0.008$
    & $45.820 \pm 0.361$ \\
Student training
    & $7.394 \pm 1.274$
    & $1645.090 \pm 286.008$ \\
Total pipeline
    & $7.585 \pm 1.264$
    & $1698.733 \pm 286.043$ \\
\hline
\multicolumn{3}{l}{\textit{Class-prediction inference}} \\
Class prediction on test set (s)
    & $0.00223 \pm 0.00013$
    & $0.66579 \pm 0.02883$ \\
Classification latency (ms/instance)
    & $0.01781 \pm 0.00106$
    & $0.04756 \pm 0.00206$ \\
\hline
\end{tabular}
\end{table*}

Training and inference also have different resource requirements. The auxiliary structures used to constrain connectivity during optimization need not be retained as separate objects at inference time, provided that the deployed representation preserves the trained model's behavior. Inference uses the trained network parameters and does not require repeating teacher training, architecture construction, or student optimization. We measured inference as the time required to predict class labels for all instances in the test set. As reported in Table~\ref{tab:sbn_training_time}, the mean class-prediction time was $0.00223 \pm 0.00013$~s for balance-scale and $0.66579 \pm 0.02883$~s for MNIST, corresponding to mean classification latencies of $0.01781 \pm 0.00106$~ms and $0.04756 \pm 0.00206$~ms per instance, respectively.

The reported runtimes should be interpreted in the context of the
hardware described in Section~\ref{subsec:methods}. The complete training
pipeline for MNIST took approximately 28.3 minutes on the NVIDIA
GeForce GTX 1080 system. These measurements characterize the training
and inference times of the evaluated SBN configurations under the
specified experimental conditions. Since the evaluation covers only
the smallest and largest benchmark configurations on a single hardware
system, the results do not establish runtime trends across all dataset
sizes or permit broad comparisons of runtime across different hardware
platforms.

\section{Discussion}

The results show that Symbolic Branch Networks provide a structured neural
representation derived from tree ensembles while retaining competitive
predictive performance on multiclass tabular datasets. The main contribution
of the approach is not the branch-to-neuron mapping itself, which has been
explored in previous tree-to-network methods, but the systematic formulation
of the resulting architecture through the structural matrices $W_1$ and
$W_2$, the calibration parameters $r_1$ and $r_2$, and the four
$\mathrm{SBN}_{a,b}$ training regimes. The results therefore provide evidence
that teacher-derived tree structure can serve as a useful structural prior for
gradient-based neural learning.

\paragraph{Predictive performance and comparison with external baselines}

Across the 29 multiclass datasets, the primary configuration
$\mathrm{SBN}_{1,0}$ achieves a mean accuracy of 87.9\%, compared with
84.02\% for the teacher ensemble. It improves upon the teacher on 28 of the
29 datasets, and the one-sided Wilcoxon signed-rank test indicates a
statistically significant improvement across the paired dataset-level
results. The comparison among the four SBN configurations also shows that
$\mathrm{SBN}_{1,0}$ achieves the highest mean accuracy, suggesting that
allowing the feature-to-branch matrix $W_1$ to adapt while retaining the
teacher-derived $W_2$ is an effective configuration within the experimental
protocol considered here.

The comparison with external baselines provides a different perspective.
$\mathrm{SBN}_{1,0}$ achieves a mean accuracy of 87.9\%, compared with
87.3\% for XGBoost and 91.3\% for TabPFN. The per-dataset results show
substantial variation in the relative performance of the three methods.
The two-sided Wilcoxon signed-rank test indicates a statistically significant
difference between $\mathrm{SBN}_{1,0}$ and TabPFN, whereas the difference
between $\mathrm{SBN}_{1,0}$ and XGBoost is not statistically significant at
the conventional 0.05 level. Thus, the higher mean accuracy of $\mathrm{SBN}_{1,0}$ 
should not be interpreted on its own as evidence of a statistically significant
difference in paired dataset-level performance.

These comparisons also highlight that predictive accuracy is only one
objective of SBN. XGBoost and TabPFN provide important predictive reference
points, but SBN is designed with an additional architectural objective:
retaining an explicit correspondence between teacher-derived branches and
neural components. The results therefore support viewing SBN as a
tree-structured neural representation rather than simply as another
predictive alternative to established tabular methods.

Results on binary classification tasks, reported separately in
\ref{app:binary}, exhibit more mixed behavior. This should not be
interpreted as a fundamental limitation of SBNs for binary classification.
Rather, the experimental protocol was designed around a shared configuration
across datasets, and further investigation of binary-specific model
configurations may be useful. Extending the evaluation to a broader range of
binary datasets and examining whether different teacher or training
configurations are beneficial in this setting remain open questions.

\paragraph{Structural interpretability}

The interpretability properties of SBN should be understood primarily in
terms of explicit structural correspondence rather than preservation of the
complete decision semantics of the teacher ensemble. Each hidden neuron
corresponds to a specific root-to-parent-of-leaf branch extracted from the
teacher ensemble. The structural matrices $W_1$ and $W_2$ encode, respectively,
the feature-to-branch and branch-to-class relationships derived from this
structure. Consequently, the architecture provides an explicit basis for
inspecting which features are associated with a branch-neuron and how that
neuron is associated with class evidence.

This correspondence is preserved structurally during training through the
fixed sparsity masks. In $\mathrm{SBN}_{1,0}$, optimization can change the
numerical values of the permitted entries of $W_1$, but it cannot introduce
new feature-to-branch connections outside the teacher-derived mask. The
learned parameters therefore refine the numerical representation within the
existing structural connectivity rather than replacing that connectivity
with an unconstrained neural representation.

An important qualification is that SBN does not preserve the original
thresholds, inequality tests, or complete decision rules of the teacher
ensemble. A hidden neuron corresponds to a branch extracted from the teacher,
but its activation in the SBN is not equivalent to evaluating whether an
input satisfies all of the original tree conditions on that branch. The
interpretability provided by SBN is therefore based on traceable structural
correspondence rather than exact reproduction of the teacher's decision
logic.

\paragraph{Global and local interpretation}

The global and local interpretability analyses provide complementary forms of
information. Global structural importance is derived from the binary
feature-to-branch mask $M_1$. The quantity
$\mathrm{GI}(i)$ measures the proportion of teacher-derived branches in which
feature $i$ occurs. It therefore characterizes the prevalence of a feature
throughout the branch-based architecture and is independent of the learned
numerical values of $W_1$.

The MNIST experiment illustrates this distinction. The structural feature
importance derived from the SBN branch connectivity exhibits a strong
Spearman rank correlation with the feature importance of the teacher
ensemble ($\rho=0.982$). This indicates that features occurring frequently
throughout the teacher-derived branch structure tend to coincide with
features receiving high importance in the teacher model. The result should
not, however, be interpreted as evidence that training preserves a learned
weighted importance profile, because the SBN global measure is determined by
the fixed structural mask rather than by the optimized values of $W_1$.

Local interpretation addresses a different question. For an individual
sample, the explanation combines the activation of selected branch-neurons
with the learned effective feature-to-branch weights and the branch-to-class
weights. The resulting contributions are sample-specific and depend on the
input, the learned parameters, and the activation state of the network.
Because each selected neuron has an associated teacher-derived branch, the
local explanation can be traced from a prediction to a specific structural
component and then to the features and class contributions associated with
that component.

This distinction is important for interpreting the role of the learned
weights. Structural feature importance describes which features are
represented throughout the teacher-derived architecture, whereas local
feature contributions describe how the learned network uses those features
for a particular input. The two measures therefore answer different
interpretability questions and should not be treated as interchangeable.

\paragraph{Relation to human understanding and symbolic structure}

The explicit branch-to-neuron correspondence provides a representation that
can be inspected at the level of individual structural components. This can
be useful in applications where users need to examine which features are
associated with particular model components or trace an individual
prediction to selected branches. However, the present experiments do not
establish that these structural explanations correspond directly to human
reasoning or domain-expert concepts.

The symbolic information used by SBN is inherited from the teacher ensemble.
In the current implementation, tree ensembles provide an automated mechanism
for extracting feature-to-branch and branch-to-class structure. The same
general framework could in principle be applied to other sources of
structured rules, provided that an appropriate branch or component
representation can be constructed. Evaluating such alternatives, especially
structures derived from domain-specific knowledge, would be a useful
direction for future work.

\paragraph{Structural constraints and neural optimization}

SBN occupies an intermediate position between a conventional tree ensemble
and an unconstrained neural network. The architecture does not learn an
arbitrary hidden topology: the number of branch-neurons and the structural
connectivity of the network are determined by the teacher ensemble. At the
same time, selected numerical parameters can be optimized through
gradient-based learning. In $\mathrm{SBN}_{1,0}$, this means that the
feature-to-branch weights can adapt while their structural support remains
fixed.

The results across the four training regimes provide evidence that this
combination of fixed structural support and numerical adaptation can be
effective. In particular, $\mathrm{SBN}_{0,0}$ achieves a mean accuracy of
86.1\% while keeping both $W_1$ and $W_2$ fixed, whereas
$\mathrm{SBN}_{1,0}$ reaches 87.9\% when $W_1$ is allowed to adapt. This
difference illustrates that the teacher-derived architecture itself carries
predictive information, while optimization of the numerical feature-to-branch
weights can provide additional predictive adaptation.

The learned scalar parameters $r_1$ and $r_2$ provide additional global
calibration of the two structural transformations. Because these parameters
and the batch-normalization layers affect the numerical scale of the
network, changes in the magnitude of $W_1$ should not be interpreted directly
as changes in structural interpretability. The structural property of
interest is instead the preservation of the connectivity pattern defined by
the fixed masks.

\subsection{Limitations and future directions}

Several limitations should be considered when interpreting the present
results. First, SBN depends on a teacher ensemble to construct its neural
architecture. The resulting branch structure, number of hidden neurons, and
feature-to-branch connectivity therefore depend on the properties of the
chosen teacher. Although the present experiments use a fixed and
dataset-independent ExtraTrees construction protocol, alternative teacher
ensembles may produce substantially different SBN architectures.

Second, the mapping retains structural information from the teacher but does
not preserve the teacher's original thresholds, inequality tests, or complete
decision rules. Consequently, the correspondence between a branch-neuron and
a teacher branch should not be interpreted as an exact neural implementation
of the original tree decision path. Future work could investigate mappings
that preserve a larger portion of the original decision semantics while
remaining compatible with differentiable optimization.

Third, the global interpretability experiment is currently demonstrated on
MNIST, while the local analysis and the explanation-stability assessment are based on a single Vehicle Silhouettes
sample. These experiments demonstrate how the proposed interpretation
mechanisms can be used, but they do not establish their usefulness across
different application domains or their stability across samples and trained
models. More extensive case studies, including repeated local explanations
and evaluations by domain experts, would provide stronger evidence regarding
the practical utility of the proposed explanations.

Fourth, the evaluation of interpretability itself remains a limitation. Quantitative metrics can 
characterize specific properties of explanations, such as stability, sparsity, 
or fidelity, but these properties do not necessarily establish whether an 
explanation is meaningful or useful in its intended application. 
In domains requiring specialized knowledge, interpretability also concerns whether 
experts can understand an explanation, relate it to the problem context, and 
use it to reason about a model's prediction. Such judgments may vary across 
experts and depend on the intended purpose of the explanation. Consequently, 
no single quantitative measure should be assumed to provide a comprehensive 
assessment of interpretability. Future work should investigate evaluation 
protocols that combine quantitative analyses with systematic assessments by 
domain experts, while explicitly defining the interpretability objectives 
and practical tasks being evaluated.

Fifth, the present study focuses on multiclass classification and uses a
shared experimental protocol across datasets. The mixed results observed in
the binary classification experiments suggest that additional investigation
of task-specific configurations may be informative. Extensions to regression,
multilabel prediction, and other structured prediction problems would also
test the generality of the approach.

Finally, the present work focuses on a systematic formulation and evaluation
of the SBN framework rather than an exhaustive comparison with every
tree-to-network architecture under its application-specific training
protocol. More extensive experiments with alternative teacher ensembles,
different tree granularities, and larger benchmark collections could clarify
how the choice of teacher structure affects both predictive performance and
the resulting interpretability properties.

Overall, the results indicate that tree-derived structure can serve as a
useful prior for neural learning without requiring the resulting model to
preserve the complete decision semantics of the original tree ensemble. SBN
combines an explicitly traceable structural architecture with gradient-based
adaptation of selected numerical parameters. This provides a basis for
studying how symbolic structure and neural optimization can be combined in
tabular learning while keeping their respective roles distinguishable.

\section{Conclusion}

This work presents the Symbolic Branch Networks (SBNs) framework, neural 
architectures whose hidden-layer structure is derived from decision-tree 
ensembles. Each root-to-parent-of-leaf branch is mapped to a hidden neuron, 
while the structural matrices $W_1$ and $W_2$ encode the feature-to-branch 
and branch-to-class relationships extracted from the teacher ensemble. The
resulting architecture preserves an explicit structural correspondence
between teacher-derived branches and neural components while allowing
selected numerical parameters to be optimized through gradient-based
learning.

Across 29 multiclass datasets, the primary configuration
$\mathrm{SBN}_{1,0}$ improves upon the teacher ensemble on 28 of 29 datasets,
increasing mean accuracy from 84.0\% to 87.9\%. The one-sided Wilcoxon
signed-rank test indicates a statistically significant improvement over the
teacher ensemble. 
Relative to external baselines, $\mathrm{SBN}_{1,0}$ achieves a mean accuracy of 87.9\%, compared with 87.3\% for XGBoost and 91.3\% for TabPFN. The corresponding two-sided Wilcoxon tests indicate a statistically significant difference between SBN$_{1,0}$ and TabPFN, whereas the difference between SBN$_{1,0}$ and XGBoost is not statistically significant at the conventional 0.05 level.

The systematic evaluation of the four SBN training regimes further shows
that $\mathrm{SBN}_{1,0}$ achieves the highest mean accuracy among the
considered SBN variants. At the same time, $\mathrm{SBN}_{0,0}$ achieves a
mean accuracy of 86.1\% while keeping both structural matrices fixed,
indicating that the teacher-derived architecture itself provides a
substantial predictive basis. Allowing $W_1$ to adapt provides additional
predictive flexibility while preserving its teacher-derived structural
connectivity through the fixed sparsity mask.

The interpretability analysis distinguishes between structural global
importance and sample-specific local explanation. Global feature importance
is derived from the fixed feature-to-branch mask $M_1$, measuring the
prevalence of each feature across the teacher-derived branches. On MNIST,
this structural measure exhibits a Spearman correlation of $\rho=0.982$ with
the feature importance of the teacher ensemble. Local explanations instead
use the learned numerical parameters and sample-specific activations to
trace predictions to selected branch-neurons, their associated features, and
their class contributions.

The local explanation stability analysis illustrates that prediction consistency and explanation consistency are distinct properties. Although the evaluated SBNs produced the same prediction for the selected instance across all runs, the features highlighted by the local explanations showed only partial overlap. This indicates that consistent predictions do not necessarily imply identical feature-based explanations, motivating the separate evaluation of prediction and explanation stability.

SBN does not preserve the original thresholds, inequality tests, or complete
decision-rule semantics of the teacher ensemble. Its symbolic character
instead lies in the explicit structural correspondence between teacher
branches and neural components and in the resulting feature-to-branch and
branch-to-class connectivity. This distinction defines both the scope and
the limitations of the proposed interpretation framework.

Several directions remain open. Future work should evaluate the stability
and practical usefulness of local explanations across larger collections of
samples and application domains, including assessment by domain experts.
Further studies could also investigate alternative teacher ensembles,
different branch-construction strategies, mappings that retain more of the
original decision semantics, and extensions to regression and other
structured prediction tasks.

Overall, the results demonstrate that tree-derived structural information
can be combined with neural optimization in a systematic manner. Symbolic
Branch Networks provide an explicit structural basis for interpreting neural
components while allowing numerical parameters to adapt to the prediction
task, offering a framework for studying the interaction between symbolic
structure and differentiable learning in tabular data.

\section*{Declaration of generative AI and AI-assisted technologies in the manuscript preparation process}
During the preparation of this work, the authors used ChatGPT (OpenAI) to assist with language refinement and to improve the clarity and organization of the manuscript. The authors reviewed and edited the suggestions and take full responsibility for the content of the published article.

\section*{Declaration of competing iterest}
The authors declare that they have no known competing financial
interests or personal relationships that could have appeared to influence
the work reported in this paper.

\bibliographystyle{elsarticle-num}
\bibliography{references}

@inproceedings{rguez2013,
  author    = {Rodr{\'i}guez-Salas, D. and G{\'o}mez-Gil, P. and Olvera-L{\'o}pez, A.},
  title     = {Designing Partially-Connected, Multilayer Perceptron Neural Nets Through Information Gain},
  booktitle = {International Joint Conference on Neural Networks (IJCNN)},
  year      = {2013},
  pages     = {1--8}, 
  publisher = {IEEE},
  doi       = {10.1109/IJCNN.2013.6706915}
}

@inproceedings{rguez2019,
  title={{ForestNet} -- Automatic Design of Sparse Multilayer Perceptron Network Architectures Using Ensembles of Randomized Trees},
  author={Rodr{\'\i}guez-Salas, Dalia and Ravikumar, Nishant and Seuret, Mathias and Maier, Andreas},
  booktitle={Asian Conference on Pattern Recognition},
  pages={32--45},
  year={2019},
  organization={Springer}
}

@article{rguez2020,
author={Rodr{\'i}guez-Salas, Dalia
and M{\"u}rschberger, Nina
and Ravikumar, Nishant
and Seuret, Mathias
and Maier, Andreas},
title={Mapping Ensembles of Trees to Sparse, Interpretable Multilayer Perceptron Networks},
journal={SN Computer Science},
year={2020},
month={Aug},
day={06},
volume={1},
number={5},
pages={252},
issn={2661-8907},
doi={10.1007/s42979-020-00268-y}
}

@article{sethi1990entropynet,
  author={Sethi, I.K.},
  journal={Proceedings of the IEEE}, 
  title={Entropy nets: from decision trees to neural networks}, 
  year={1990},
  volume={78},
  number={10},
  pages={1605-1613},
  doi={10.1109/5.58346}}

@inproceedings{kontschieder2015deep,
  author={Kontschieder, Peter and Fiterau, Madalina and Criminisi, Antonio and Bulò, Samuel Rota},
  booktitle={2015 IEEE International Conference on Computer Vision (ICCV)}, 
  title={Deep Neural Decision Forests}, 
  year={2015},
  volume={},
  number={},
  pages={1467-1475},
  doi={10.1109/ICCV.2015.172}}

@inproceedings{perez2023,
author={Pérez-Toro, P. A. and Rodr{\'i}guez-Salas, D. and Arias-Vergara, T. and Bayerl, S. P. and Klumpp, P. and Riedhammer, K. and Schuster, M. and Nöth, E. and Maier, A. and Orozco-Arroyave, J. R.},
booktitle={ICASSP 2023 - 2023 IEEE International Conference on Acoustics, Speech and Signal Processing (ICASSP)}, 
title={Transferring Quantified Emotion Knowledge for the Detection of Depression in {Alzheimer}’s Disease Using {ForestNets}}, 
year={2023},
pages={1-5},
doi={10.1109/ICASSP49357.2023.10095219}
}

@article{perez2022,
title = {Interpreting Acoustic Features for the Assessment of {Alzheimer}’s Disease using {ForestNet}},
journal = {Smart Health},
volume = {26},
pages = {100347},
year = {2022},
issn = {2352-6483},
doi = {10.1016/j.smhl.2022.100347},
author = {Pérez-Toro, P. A. and Rodr{\'i}guez-Salas, D. and Arias-Vergara, T. 
          and Klumpp, P. 
          and Schuster, M. 
          and Nöth, E. 
          and Orozco-Arroyave, J. R. 
          and Maier, A. K.}
}

@inproceedings{treevsnn,
author = {McElfresh, Duncan and Khandagale, Sujay and Valverde, Jonathan and C., Vishak Prasad and Ramakrishnan, Ganesh and Goldblum, Micah and White, Colin},
title = {When do neural nets outperform boosted trees on tabular data?},
year = {2023},
publisher = {Curran Associates Inc.},
address = {Red Hook, NY, USA},
articleno = {3337},
numpages = {34},
booktitle = {Proceedings of the 37th International Conference on Neural Information Processing Systems},
pages={76336 - 76369},
location = {New Orleans, LA, USA},
series = {NIPS '23}
}

@misc{my-arxiv,
      title={BranchNet: A Neuro-Symbolic Learning Framework for Structured Multi-Class Classification}, 
      author={Dalia Rodríguez-Salas and Christian Riess},
      year={2025},
      eprint={2507.01781},
      archivePrefix={arXiv},
      primaryClass={cs.LG},
      url={https://arxiv.org/abs/2507.01781v1}, 
}

@inproceedings{chen2016xgboost,
  title     = {XGBoost: A Scalable Tree Boosting System},
  author    = {Chen, Tianqi and Guestrin, Carlos},
  booktitle = {Proceedings of the 22nd ACM SIGKDD International Conference on Knowledge Discovery and Data Mining},
  year      = {2016},
  pages     = {785--794},
  publisher = {ACM}
}

@article{xiao2017nrf,
author={Biau, G{\'e}rard
and Scornet, Erwan
and Welbl, Johannes},
title={Neural Random Forests},
journal={Sankhya A},
year={2019},
month={Dec},
day={01},
volume={81},
number={2},
pages={347-386},
issn={0976-8378},
doi={10.1007/s13171-018-0133-y},
url={https://doi.org/10.1007/s13171-018-0133-y}
}

@inproceedings{irsoy2012soft,
  author={İrsoy, Ozan and Yıldız, Olcay Taner and Alpaydın, Ethem},
  booktitle={Proceedings of the 21st International Conference on Pattern Recognition (ICPR2012)}, 
  title={Soft decision trees}, 
  year={2012},
  volume={},
  number={},
  pages={1819-1822},
  doi={}}

@inproceedings{rguez2024,
  author="Rodr{\'i}guez-Salas, Dalia
    and Riess, Christian
    and Vicario, Celia Mart{\'i}n
    and Taubmann, Oliver
    and Ditt, Hendrik
    and Schwab, Stefan
    and D{\"o}rfler, Arnd",
  title="Analysing Variables for 90-Day Functional-Outcome Prediction of Endovascular Thrombectomy",
  booktitle="Medical Image Understanding and Analysis",
  year="2024",
  publisher="Springer Nature Switzerland",
  address="Cham",
  pages="202--215",
  doi       = {10.1007/978-3-031-61019-3_35} 
}

@inproceedings{da23-1,
  author={Rodríguez-Salas, Dalia and Öttl, Mathias and Seuret, Mathias and Packhäuser, Kai and Maier, Andreas},
  booktitle={2023 IEEE 20th International Symposium on Biomedical Imaging (ISBI)}, 
  title={Using {Forestnets} for Partial Fine-Tuning Prior to Breast Cancer Detection in Ultrasounds}, 
  year={2023},
  volume={},
  number={},
  pages={1-5},
  doi={10.1109/ISBI53787.2023.10230424}}

@article{FRIEDMAN2002367,
title = {Stochastic gradient boosting},
journal = {Computational Statistics \& Data Analysis},
volume = {38},
number = {4},
pages = {367-378},
year = {2002},
note = {Nonlinear Methods and Data Mining},
issn = {0167-9473},
doi = {https://doi.org/10.1016/S0167-9473(01)00065-2},
url = {https://www.sciencedirect.com/science/article/pii/S0167947301000652},
author = {Jerome H. Friedman}
}

@Article{tabpfn,
author={Hollmann, Noah
and M{\"u}ller, Samuel
and Purucker, Lennart
and Krishnakumar, Arjun
and K{\"o}rfer, Max
and Hoo, Shi Bin
and Schirrmeister, Robin Tibor
and Hutter, Frank},
title={Accurate predictions on small data with a tabular foundation model},
journal={Nature},
year={2025},
month={Jan},
day={01},
volume={637},
number={8045},
pages={319-326},
issn={1476-4687},
doi={10.1038/s41586-024-08328-6},
url={https://doi.org/10.1038/s41586-024-08328-6}
}

@Inbook{molnar2022,
author="Molnar, Christoph
and K{\"o}nig, Gunnar
and Herbinger, Julia
and Freiesleben, Timo
and Dandl, Susanne
and Scholbeck, Christian A.
and Casalicchio, Giuseppe
and Grosse-Wentrup, Moritz
and Bischl, Bernd",
title="General Pitfalls of Model-Agnostic Interpretation Methods for Machine Learning Models",
bookTitle="xxAI - Beyond Explainable AI: International Workshop, Held in Conjunction with ICML 2020, July 18, 2020, Vienna, Austria, Revised and Extended Papers",
year="2022",
publisher="Springer International Publishing",
address="Cham",
chapter="4",
pages="39--68",
isbn="978-3-031-04083-2",
doi="10.1007/978-3-031-04083-2_4",
url="https://doi.org/10.1007/978-3-031-04083-2_4"
}

@article{zschech2026,
author={Zschech, Patrick
and Weinzierl, Sven
and Kraus, Mathias},
title={Inherently Interpretable Machine Learning: A Contrasting Paradigm to Post-hoc Explainable AI},
journal={Business {\&} Information Systems Engineering},
year={2026},
month={Apr},
day={01},
volume={68},
number={2},
pages={445-463},
issn={1867-0202},
doi={10.1007/s12599-025-00964-0},
url={https://doi.org/10.1007/s12599-025-00964-0}
}

@article{oml-benchmarking-suites,
      title={OpenML Benchmarking Suites}, 
      author={Bernd Bischl and Giuseppe Casalicchio and Matthias Feurer and Frank Hutter and Michel Lang and Rafael G. Mantovani and Jan N. van Rijn and Joaquin Vanschoren},
      year={2019},
      journal={arXiv:1708.03731v2 [stat.ML]}
}

@article{OpenML2025,
      author = {Bernd Bischl and Giuseppe Casalicchio and Taniya Das and Matthias Feurer and Sebastian Fischer and Pieter Gijsbers and Subhaditya Mukherjee and Andreas C Müller and László Németh and Luis Oala and Lennart Purucker and Sahithya Ravi and Jan N van Rijn and Prabhant Singh and Joaquin Vanschoren and Jos van der Velde and Marcel Wever},
      title = {OpenML: Insights from 10 years and more than a thousand papers},
      journal = {Patterns},
      volume = {6},
      number = {7},
      year = {2025},
      pages = {101317},
      url = {https://www.cell.com/patterns/fulltext/S2666-3899(25)00165-5},
      doi = {10.1016/j.patter.2025.101317},
      publisher = {Cell Press}
    }


\appendix

\section{Binary Classification Results}
\label{app:binary}

The main experiments focus on multiclass classification, consistent with the
intended scope of the proposed SBN framework. For completeness, this appendix
reports the corresponding results on the binary classification datasets in the
OpenML-CC18 collection. The same experimental protocol is used as in the main
experiments: for each dataset, the data are evaluated over 10 random seeds,
with a $70\%/10\%/20\%$ train/validation/test split. The validation set is used
for early stopping and learning-rate scheduling, while the test set is used
only for the final evaluation.

For binary classification, performance is evaluated using the area under the
receiver operating characteristic curve (AUC). For each dataset and model, the
reported value is the mean AUC across the 10 random seeds, with the standard
deviation shown after the $\pm$ symbol.

The binary experiments include the teacher ensemble, denoted
\textbf{TeacherSBN}, and the four SBN training configurations
$\mathrm{SBN}_{0,0}$, $\mathrm{SBN}_{0,1}$, $\mathrm{SBN}_{1,0}$, and
$\mathrm{SBN}_{1,1}$. As in the main experiments, the notation specifies
whether the teacher-derived feature-to-branch matrix $W_1$ and
branch-to-class matrix $W_2$ are trainable.

\subsection{Performance on Binary Classification Tasks}
\label{subsec:binary_performance}

Table~\ref{tab:auc_binary} reports the performance of TeacherSBN and the four
SBN configurations across the binary datasets. The results show more
heterogeneous behavior than that observed in the multiclass benchmark. While
several datasets exhibit substantial improvements over TeacherSBN, other
datasets show smaller gains or reductions in performance depending on the SBN
configuration.

\begin{table*}
\centering
\caption{Mean AUC across 10 random seeds for each SBN variant and their teacher
ensemble, TeacherSBN, across 30 binary datasets. Values
are reported as mean $\pm$ standard deviation.}
\begin{tabular}{lccccc}
\toprule
Dataset & TeacherSBN & SBN\_0\_0 & SBN\_0\_1 & SBN\_1\_0 & SBN\_1\_1 \\
\midrule
Internet-Advertisements & 0.950 $\pm$ 0.011 & 0.931 $\pm$ 0.018 & 0.933 $\pm$ 0.019 & 0.921 $\pm$ 0.023 & 0.922 $\pm$ 0.023 \\
PhishingWebsites & 0.959 $\pm$ 0.003 & 0.957 $\pm$ 0.003 & 0.955 $\pm$ 0.003 & 0.971 $\pm$ 0.003 & 0.970 $\pm$ 0.004 \\
adult & 0.808 $\pm$ 0.004 & 0.820 $\pm$ 0.005 & 0.819 $\pm$ 0.003 & 0.803 $\pm$ 0.006 & 0.791 $\pm$ 0.046 \\
bank-marketing & 0.819 $\pm$ 0.013 & 0.861 $\pm$ 0.005 & 0.860 $\pm$ 0.005 & 0.839 $\pm$ 0.008 & 0.846 $\pm$ 0.006 \\
banknote-authentication & 0.993 $\pm$ 0.005 & 1.000 $\pm$ 0.000 & 0.999 $\pm$ 0.002 & 1.000 $\pm$ 0.000 & 1.000 $\pm$ 0.000 \\
blood-transfusion-service & 0.625 $\pm$ 0.053 & 0.685 $\pm$ 0.034 & 0.685 $\pm$ 0.033 & 0.685 $\pm$ 0.036 & 0.683 $\pm$ 0.031 \\
breast-w & 0.968 $\pm$ 0.014 & 0.969 $\pm$ 0.016 & 0.961 $\pm$ 0.015 & 0.965 $\pm$ 0.017 & 0.963 $\pm$ 0.016 \\
churn & 0.804 $\pm$ 0.031 & 0.864 $\pm$ 0.023 & 0.859 $\pm$ 0.031 & 0.835 $\pm$ 0.023 & 0.829 $\pm$ 0.016 \\
climate-model-simulation & 0.763 $\pm$ 0.074 & 0.858 $\pm$ 0.076 & 0.847 $\pm$ 0.071 & 0.852 $\pm$ 0.074 & 0.882 $\pm$ 0.056 \\
credit-g & 0.658 $\pm$ 0.030 & 0.663 $\pm$ 0.039 & 0.635 $\pm$ 0.034 & 0.666 $\pm$ 0.034 & 0.673 $\pm$ 0.025 \\
cylinder-bands & 0.769 $\pm$ 0.039 & 0.764 $\pm$ 0.034 & 0.761 $\pm$ 0.030 & 0.805 $\pm$ 0.040 & 0.793 $\pm$ 0.025 \\
diabetes & 0.712 $\pm$ 0.031 & 0.744 $\pm$ 0.030 & 0.727 $\pm$ 0.031 & 0.724 $\pm$ 0.030 & 0.726 $\pm$ 0.037 \\
dresses-sales & 0.559 $\pm$ 0.052 & 0.532 $\pm$ 0.063 & 0.532 $\pm$ 0.046 & 0.549 $\pm$ 0.036 & 0.550 $\pm$ 0.040 \\
electricity & 0.755 $\pm$ 0.011 & 0.774 $\pm$ 0.010 & 0.776 $\pm$ 0.006 & 0.800 $\pm$ 0.006 & 0.787 $\pm$ 0.006 \\
ilpd & 0.666 $\pm$ 0.039 & 0.667 $\pm$ 0.048 & 0.671 $\pm$ 0.055 & 0.655 $\pm$ 0.038 & 0.618 $\pm$ 0.062 \\
jm1 & 0.658 $\pm$ 0.008 & 0.658 $\pm$ 0.014 & 0.655 $\pm$ 0.009 & 0.637 $\pm$ 0.034 & 0.648 $\pm$ 0.020 \\
kc1 & 0.734 $\pm$ 0.031 & 0.712 $\pm$ 0.028 & 0.716 $\pm$ 0.031 & 0.701 $\pm$ 0.040 & 0.705 $\pm$ 0.033 \\
kc2 & 0.796 $\pm$ 0.039 & 0.769 $\pm$ 0.041 & 0.765 $\pm$ 0.050 & 0.762 $\pm$ 0.063 & 0.713 $\pm$ 0.073 \\
kr-vs-kp & 0.986 $\pm$ 0.005 & 0.990 $\pm$ 0.003 & 0.992 $\pm$ 0.004 & 0.993 $\pm$ 0.004 & 0.993 $\pm$ 0.004 \\
madelon & 0.588 $\pm$ 0.018 & 0.657 $\pm$ 0.028 & 0.672 $\pm$ 0.023 & 0.658 $\pm$ 0.022 & 0.647 $\pm$ 0.017 \\
numerai28 & 0.515 $\pm$ 0.003 & 0.518 $\pm$ 0.004 & 0.519 $\pm$ 0.004 & 0.515 $\pm$ 0.003 & 0.519 $\pm$ 0.002 \\
ozone-level-8hr & 0.794 $\pm$ 0.026 & 0.823 $\pm$ 0.024 & 0.828 $\pm$ 0.028 & 0.806 $\pm$ 0.025 & 0.806 $\pm$ 0.036 \\
pc1 & 0.726 $\pm$ 0.033 & 0.795 $\pm$ 0.038 & 0.791 $\pm$ 0.049 & 0.774 $\pm$ 0.047 & 0.773 $\pm$ 0.038 \\
pc3 & 0.746 $\pm$ 0.035 & 0.750 $\pm$ 0.054 & 0.732 $\pm$ 0.062 & 0.664 $\pm$ 0.062 & 0.677 $\pm$ 0.078 \\
phoneme & 0.817 $\pm$ 0.016 & 0.822 $\pm$ 0.011 & 0.826 $\pm$ 0.009 & 0.857 $\pm$ 0.011 & 0.853 $\pm$ 0.009 \\
qsar-biodeg & 0.836 $\pm$ 0.022 & 0.844 $\pm$ 0.014 & 0.840 $\pm$ 0.018 & 0.851 $\pm$ 0.011 & 0.848 $\pm$ 0.021 \\
spambase & 0.938 $\pm$ 0.010 & 0.934 $\pm$ 0.010 & 0.933 $\pm$ 0.009 & 0.938 $\pm$ 0.007 & 0.932 $\pm$ 0.009 \\
tic-tac-toe & 0.918 $\pm$ 0.021 & 0.981 $\pm$ 0.011 & 0.974 $\pm$ 0.014 & 0.997 $\pm$ 0.006 & 0.997 $\pm$ 0.008 \\
wdbc & 0.951 $\pm$ 0.018 & 0.967 $\pm$ 0.025 & 0.954 $\pm$ 0.022 & 0.972 $\pm$ 0.019 & 0.951 $\pm$ 0.025 \\
wilt & 0.882 $\pm$ 0.046 & 0.973 $\pm$ 0.013 & 0.970 $\pm$ 0.011 & 0.980 $\pm$ 0.007 & 0.979 $\pm$ 0.010 \\
\bottomrule
\end{tabular}
\label{tab:auc_binary}
\end{table*}

\subsection{Statistical Comparison with TeacherSBN}
\label{subsec:binary_statistics}

To assess whether the SBN configurations consistently improve upon the teacher
ensemble across binary datasets, we perform one-sided Wilcoxon signed-rank tests
on the paired dataset-level results. For each dataset, the mean test performance
across the 10 random seeds is treated as a single paired observation. The
directional alternative corresponds to higher performance for the SBN
configuration than for TeacherSBN.

\begin{table*}[ht]
\centering
\small
\caption{One-sided Wilcoxon signed-rank tests comparing the AUC of each SBN variant against their teacher ensemble, TeacherSBN, across the binary classification datasets.}
\label{tab:wilcoxon_binary}
\begin{tabular}{lcc}
\toprule
Comparison & Wilcoxon statistic & $p$-value \\
\midrule
$\mathrm{SBN}_{0,0}$ vs. TeacherSBN & 368 & $2.1691\times10^{-3}$ \\
$\mathrm{SBN}_{0,1}$ vs. TeacherSBN & 323 & $3.1778\times10^{-2}$ \\
$\mathrm{SBN}_{1,0}$ vs. TeacherSBN & 341 & $1.2393\times10^{-2}$ \\
$\mathrm{SBN}_{1,1}$ vs. TeacherSBN & 314 & $4.8051\times10^{-2}$ \\
\bottomrule
\end{tabular}
\end{table*}

The statistical results indicate that the SBN configurations exhibit positive
paired differences relative to TeacherSBN across the binary datasets, although
the strength of the evidence varies substantially between configurations. 
These results contrast with the multiclass experiments, where all four SBN
configurations produce substantially smaller one-sided Wilcoxon $p$-values.
The binary results therefore provide a complementary assessment of the
framework and indicate that the magnitude and consistency of the improvement
depend on the classification setting and SBN training regime.

\subsection{Binary Architectural Details}

Table~\ref{tab:sbn_binary} summarizes the architectural characteristics of
the SBN representations constructed for the binary classification datasets.
In addition to the number of samples and input features, the table reports
the range of hidden neurons induced by the teacher-derived branches and the
corresponding range of sparsity in the feature-to-branch matrix $W_1$. The sparsity of $W_2$ is not reported because no zero entries were observed in this matrix across the binary datasets under the tree-depth
restrictions adopted in our experiments. However, this does not imply that $W_2$ must always be dense, as unrestricted tree depth may produce branches containing samples from only one class.

The number of hidden neurons varies substantially across datasets. This
variation follows from the teacher-ensemble construction, since each
root-to-parent-of-leaf branch produces one hidden neuron. Consequently,
datasets with different numbers of features and different teacher-tree
structures produce SBNs with different hidden-layer sizes. For example,
\texttt{banknote-authentication} produces between 155 and 184 hidden neurons, whereas \texttt{adult} produces between 8424 and 8505 hidden neurons.

The sparsity of $W_1$ also varies considerably across binary datasets. For
example, \texttt{wilt} has a $W_1$ sparsity between 9.15\% and 14.38\%,
whereas high-dimensional datasets such as \texttt{madelon} and
\texttt{Internet-Advertisements} exhibit sparsity
above 96\%. This variation reflects the relationship between feature
dimensionality and the features used along the teacher-derived branches.
Because $W_1$ contains nonzero entries only for features occurring in the
corresponding branches, datasets with many input features relative to the
number of features used by individual branches can produce highly sparse
matrices.

The branch-to-class matrix $W_2$ exhibits a different structural behavior in
the binary setting. No zero entries were observed in $W_2$ across the evaluated 
binary datasets under the tree-depth
restrictions adopted in our experiments. This indicates that, in the configurations 
considered, branches generally retain connections to both
output classes. However, this behavior is not an inherent property of
binary SBNs. If tree depth is unrestricted, a branch may reach a node
containing samples from only one class, potentially resulting in zero
entries in the corresponding class connections.

This behavior follows from the construction of $W_2$ rather than from an
explicit binary-specific sparsity constraint. The matrix stores class
evidence associated with the samples reaching each parent-of-leaf node. 
Consequently, when a node contains samples from both classes, its corresponding
branch can contribute to both output classes. The observed density of
$W_2$ in our binary experiments therefore reflects the tree-depth restrictions 
and the resulting distribution of samples across the teacher-derived
branches.

The architectural complexity is also controlled by the teacher construction
protocol. The number of trees and their leaf budgets are determined from the
dataset dimensionality and number of classes, with the maximum number of leaf
nodes varied across the sequentially constructed trees. Increasing the leaf
budget produces more detailed tree partitions and can consequently increase
the number of extracted branches and hidden neurons. This creates a
trade-off between the granularity of the teacher-derived structural
representation and the resulting neural architecture size.

The binary results therefore illustrate that the architecture induced by SBN
can vary substantially across datasets. Differences in hidden-layer size,
feature-to-branch sparsity, and branch-to-class structure arise directly
from the teacher ensemble and are not imposed by a fixed neural architecture.
Further investigation of how these architectural properties interact with
binary predictive performance, including alternative tree-construction
strategies and binary-specific configurations, remains an interesting
direction for future work.

\begin{table*}[htbp]
\centering
\small
\caption{Summary of SBN architectural characteristics for binary classification datasets.}
\begin{tabular}{rrrrrrr}
\toprule
Dataset & feats & samples & min & max & min & max \\
       &       &         & \multicolumn{2}{c}{Hidden neurons} & \multicolumn{2}{c}{$W_1$ sparsity (\%)} \\
\midrule
Internet-Advertisements & 1558 & 3279 & 2858 & 3552 & 96.56 & 97.11 \\
PhishingWebsites & 30 & 11055 & 1776 & 1920 & 58.86 & 61.30 \\
adult & 108 & 48842 & 8424 & 8505 & 79.91 & 81.89 \\
bank-marketing & 51 & 45211 & 5152 & 5215 & 66.92 & 69.74 \\
banknote-authentication & 4 & 1372 & 155 & 184 & 13.04 & 20.60 \\
blood-transfusion-service & 4 & 748 & 257 & 286 & 4.68 & 12.65 \\
breast-w & 9 & 699 & 160 & 196 & 41.39 & 50.64 \\
churn & 20 & 5000 & 1028 & 1109 & 49.63 & 54.26 \\
climate-model-simulation & 18 & 540 & 230 & 285 & 65.36 & 68.63 \\
credit-g & 61 & 1000 & 1134 & 1193 & 81.13 & 82.68 \\
cylinder-bands & 131 & 540 & 833 & 946 & 90.38 & 91.98 \\
diabetes & 8 & 768 & 502 & 520 & 24.03 & 29.13 \\
dresses-sales & 160 & 500 & 1058 & 1122 & 90.00 & 91.95 \\
electricity & 8 & 45312 & 488 & 505 & 22.11 & 27.55 \\
ilpd & 11 & 583 & 407 & 435 & 31.35 & 40.58 \\
jm1 & 21 & 10885 & 1269 & 1292 & 43.58 & 51.84 \\
kc1 & 21 & 2109 & 772 & 812 & 55.30 & 59.16 \\
kc2 & 21 & 522 & 355 & 392 & 62.61 & 66.51 \\
kr-vs-kp & 73 & 3196 & 678 & 1041 & 82.93 & 84.81 \\
madelon & 500 & 2600 & 2620 & 2717 & 97.18 & 97.47 \\
numerai28 & 21 & 96320 & 1237 & 1258 & 52.82 & 55.10 \\
ozone-level-8hr & 72 & 2534 & 794 & 894 & 85.87 & 87.07 \\
pc1 & 21 & 1109 & 378 & 441 & 60.58 & 63.83 \\
pc3 & 37 & 1563 & 720 & 798 & 74.22 & 76.45 \\
phoneme & 5 & 5404 & 306 & 317 & 9.17 & 16.10 \\
qsar-biodeg & 41 & 1055 & 755 & 810 & 73.12 & 76.82 \\
spambase & 57 & 4601 & 2106 & 2355 & 64.63 & 71.43 \\
tic-tac-toe & 27 & 958 & 592 & 685 & 69.35 & 70.57 \\
wdbc & 30 & 569 & 213 & 260 & 77.29 & 79.89 \\
wilt & 5 & 4839 & 310 & 331 & 9.15 & 14.38 \\
\bottomrule
\end{tabular}
\label{tab:sbn_binary}
\end{table*}

\section*{Data availability}
All datasets in this study are publicly available through OpenML \cite{OpenML2025}. The implementation used in this work is publicly available at
\url{https://github.com/noeth-rodriguez/SymbolicBranchNetworks} to support reproducibility.

\end{document}